\documentclass[11pt,letterpaper]{article}

\usepackage[utf8]{inputenc}
\usepackage[T1]{fontenc}
\usepackage[english]{babel}
\usepackage{amsmath,amssymb}
\usepackage{newpxtext,newpxmath}
\usepackage[varqu,varl,scaled=0.95]{zi4}
\usepackage[margin=1in]{geometry}
\usepackage{microtype}
\usepackage{graphicx}
\usepackage{xcolor}
\usepackage{booktabs,array,longtable}
\usepackage{algorithm}
\usepackage[noend]{algpseudocode}
\usepackage[labelfont=bf,labelsep=period,font=small]{caption}
\usepackage{authblk}
\usepackage{titlesec}
\usepackage{needspace}
\usepackage[authoryear,round]{natbib}
\setcitestyle{semicolon}
\usepackage{url}
\usepackage[colorlinks=true,allcolors=blue]{hyperref}

\titleformat{\paragraph}[runin]{\normalfont\normalsize\bfseries}{\theparagraph}{1em}{}[.]

\DeclareUnicodeCharacter{2013}{\TextOrMath{\textendash}{\text{\textendash}}}

\date{}

\newcommand{\backmatter}[1]{\par\medskip\noindent\textbf{#1.}\hspace{0.5em}}

\usepackage{listings}
\newcommand{\promptfont}{\ttfamily}

\lstdefinestyle{prompt}{
  basicstyle=\promptfont\small,
  breaklines=true,
  breakatwhitespace=false,
  columns=fixed,
  keepspaces=true,
  showstringspaces=false,
  breakautoindent=false,
  breakindent=0pt,
  postbreak=\mbox{\textcolor{gray}{$\hookrightarrow$}\space},
  frame=single,
  rulecolor=\color{gray!40},
  xleftmargin=6pt,
  xrightmargin=6pt,
}

\begin{document}

\title{Automating the science of mind using agent-driven theory discovery and experimentation}

\author[a,b,1]{\mbox{Ben Prystawski}}
\author[a,*]{\mbox{Kushin Mukherjee}}
\author[a,*]{\mbox{Daniel Wurgaft}}
\author[a]{\mbox{Linas Nasvytis}}
\author[c]{\mbox{Michael Y. Li}}
\author[a,c]{\mbox{Noah D. Goodman}}
\author[a]{\mbox{Michael C. Frank}}
\affil[a]{Department of Psychology, Stanford University}
\affil[b]{Data and Intelligent Systems, Princeton University}
\affil[c]{Department of Computer Science, Stanford University}

\maketitle
\begin{NoHyper}\let\thefootnote\relax\footnotetext{\textsuperscript{1}To whom correspondence should be addressed. e-mail: benpry@princeton.edu, \textsuperscript{*}Equal contribution}\end{NoHyper}

\begin{abstract}
AI-based scientific automation is increasingly possible using agents to generate hypotheses, design experiments, and analyze data. Collecting new datasets is a major bottleneck in this pipeline, however. Psychology, and computational cognitive science in particular, is well-positioned to benefit from AI experimentation because theories are often represented as code and crowdsourcing platforms enable programmatic human data collection at scale. Here, we apply automated discovery techniques to the project of generating theories in computational cognitive science, with an agent-based system collecting human data independently through crowdsourced survey experiments. Our system, {\sc auto-psych}, uses nested agent-based discovery loops to generate explanatory theories of human behavior.  The inner loop conjectures, fits, and critiques probabilistic cognitive models; the outer loop designs experiments to test these models, launches them online, and analyzes the data. We applied {\sc auto-psych} to three classic domains in cognitive psychology: subjective randomness, reference games, and the number game. We first tested {\sc auto-psych}'s ability to recover a known theory when data is generated from it and found strong recovery accuracy. We then deployed {\sc auto-psych} in a live setting, where it ran real human experiments online. Across all three domains, the system finds theories that predict human behavior better than leading theories from the scientific literature. This work thus demonstrates the feasibility of automated data collection and theory discovery in computational cognitive science.
\end{abstract}

\begin{center}
\begin{minipage}{0.9\textwidth}
\small
\begin{center}\textbf{Significance}\end{center}
AI agents are increasingly capable of performing tasks in scientific workflows. But to date there is limited work evaluating them in the full, end-to-end process of scientific research. Cognitive psychology is an ideal test case for automation of the full loop of scientific discovery as experiments are typically run online and theories are represented as programs. We develop a system, {\sc auto-psych}, that automatically develops theories and deploys experiments to test them. {\sc auto-psych} discovers theories that predict human behavior better than leading theories from the literature in three classic domains of cognitive psychology.
\end{minipage}
\end{center}

\medskip
\noindent\textbf{Keywords:} Cognitive science $|$ Scientific automation $|$ Agentic AI $|$ Computational modeling

\bigskip

\section*{Introduction}

The scientific process is often described as an iterative loop in which researchers conjecture a theory, devise specific experimental tests of that theory, and then adjust their theory in response to evidence \citep{godfrey2009theory}.
AI tools built on large language models (LLMs)  can already assist with individual steps in this loop -- proposing theories, designing and conducting experiments, analyzing data, and providing interpretations of results \citep{li2024automatedstatisticalmodeldiscovery, gandhi2025boxing}. 
More recent frameworks have begun integrating multiple stages of the research process into more autonomous workflows \citep{jagadish2026toward, schmidgall2025agent, murphy2026model}, contributing to new results in fields like mathematics and theoretical physics \citep{davies2021advancing, romera2024mathematical, openai2026navierstokes, alpoge2026blowup, vonhippel2026nineloops}.
Could AI systems complete every part of the discovery loop? Such an advance could lead to increasing automation of the scientific process and faster progress towards important applications \citep{musslick2025automating, jagadish2026toward}. 
Further, agent-based systems can now use computational tools and integrate with data collection platforms, making this goal increasingly feasible. As a step towards it, the current paper develops a framework that automates the complete model discovery loop in computational cognitive science.

Much of the recent progress in agent-based workflows across the sciences has focused on discovering scientific models from existing datasets.
The pattern of iterative model testing and refinement on a known dataset is sometimes known as ``Box's loop'' (after statistician George Box) \citep{blei2014build}. Because no new data are needed, language 
model agents can quickly propose, fit, and critique models  \citep{li2024automatedstatisticalmodeldiscovery,li2024criticalcriticautomationlanguage, gandhi2025boxing}. 
Some methods have also combined model discovery with more active experimental design, including in simulated scientific environments \citep{eltetHo2026atlas,murphy2026model,jagadish2026toward}. However, a critical part of the scientific process is collecting \textit{new, real-world}
data under targeted experimental regimes to test theories.
Thus, extending AI-driven discovery to real-world experimentation requires a ``lab-in-the-loop'' model \citep{swanson2025virtual,ghareeb2026multi}, in which an agent-based workflow automatically designs experiments, deploys them, and analyzes data generated from those experiments.

\begin{figure}
    \centering
\includegraphics[width=\textwidth]{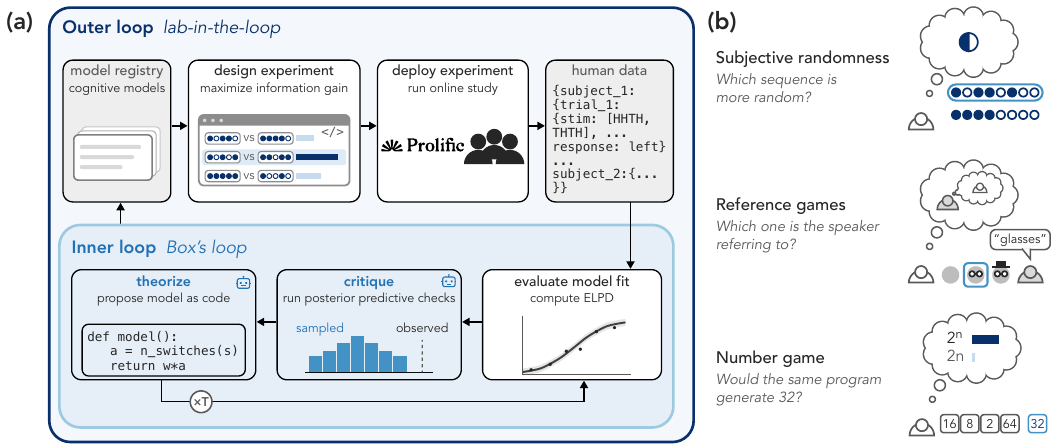}
    \caption{An agent-based discovery loop for automated cognitive science. \textbf{(a)}: In the \emph{outer loop}, stimuli are chosen using their expected information gain over a \emph{model registry}. Selected stimuli are added to an experiment and deployed to participants on Prolific. The collected human data are passed to the \emph{inner loop}. In the inner loop, each registry model is first fit to the data. A critic agent then performs posterior-predictive checks: it both proposes and computes test statistics on datasets simulated from the best incumbent model and compares this reference distribution to the value computed on the human data. A theorist agent uses this critique to propose new probabilistic models. The best-fitting models are added to the model registry. \textbf{(b)}: Illustration of the three domains we applied {\sc auto-psych} to.}
    \label{fig:loops}
\end{figure}

This lab-in-the-loop model has been achieved in machine learning research, where AI agents can run computational experiments without any humans involved. For example, one automated system for computational research decomposes each problem into a set of steps -- idea generation, novelty checking, experimentation, and paper writing \citep{lu2026towards}; parallel passes through this workflow with comparison and pruning at each step resulted in an agent-generated paper that passed human review \citep[cf.][]{beel2025evaluating}. Even for {\it in silico} systems like this, the time it takes to deploy experiments and get results is a major bottleneck to rapid iteration \citep{karpathy_autoresearch_2026}.

Much of the development in this area has focused on autonomous computational discovery, distinguishing single-agent systems \citep{karpathy_autoresearch_2026, jiang2025aide} from multi-agent systems that coordinate ensembles of agents with different roles \citep{gao2026autoscientist, swanson2025virtual} (an approach that we follow here). 
Work in both of these traditions has developed methods for automatically generating, critiquing, and revising statistical models \citep{li2024automatedstatisticalmodeldiscovery,li2024criticalcriticautomationlanguage,eltetHo2026atlas,agarwal2025autodiscovery} as well as benchmarking their experiment planning abilities \citep{kon2025exp}. 

Similar approaches have yielded positive results in other fields. For example, in structural biology, simulation tools allow for prediction of protein structures and biochemical interactions on par with \textit{in vitro} experimentation, enabling the rapid development of novel theories \citep{gottweis2026accelerating,ghareeb2026multi,swanson2025virtual, jumper2021highly, m2024augmenting, jin2025stella, gao2024empowering}. Agent-based workflows have also found solutions to open problems and proposed new algorithms in mathematics \citep{novikov2025alphaevolve, romera2024mathematical, fawzi2022discovering,openai2026navierstokes,alpoge2026blowup}, and have been applied successfully to the social sciences and cognitive science \citep{balla2025ai,rmus2025generating, jagadish2026toward,geng2025large, eltetHo2026atlas}.
Of this work, \citet{rmus2025generating} is closest to our current study as they use AI models as part of a pipeline to propose and validate probabilistic cognitive models. Thus, automating the entire process of computational cognitive science, from conjecturing and evaluating hypotheses to designing and running experiments with real human participants, is an open challenge [see \citet{jagadish2026closing} and \citet{wehr2026closing} for concurrent, complementary approaches to this challenge].

Experimentation in psychology requires human participants, but many studies are run online via crowdsourcing platforms like Amazon Mechanical Turk and Prolific \citep{buhrmester2018evaluation,palan2018prolific}. 
Such experiments are typically served in template-based web frameworks  \citep{de2015jspsych}, meaning that they can be created by agents and deployed via application programming interfaces (APIs). Just as autonomous machine learning research depends on small-scale training runs, autonomous psychology can make use of fast and inexpensive online behavioral experiments. 

Computational cognitive science is the sub-field of psychology that might have the most to gain from automated experimentation. Extensive work has been done to formalize theories in this domain. 
These theories can be expressed in a number of highly expressive and well-documented probabilistic programming languages (e.g., Stan, PyMC, Pyro) \citep{goodman2013principles,bingham2019pyro,carpenter2017stan,patil2010pymc}, meaning that they can be generated by coding agents and compared with one another using standard statistical approaches. While previous work has automatically generated and compared cognitive models of this type \citep{rmus2025generating}, their workflows have yet to be integrated into a full human experimentation loop.

Evaluating the quality of a system for automated scientific discovery is challenging. Since we do not know a ``ground truth'' model of human cognition, we cannot measure how close a discovered model is to the truth. However, a clear measurable desideratum is the ability to recover a known ground truth model. In cases where we know the ground-truth model that generated data, our model discovery process should be able to find something close to that ground truth. We can evaluate model recovery both in the distribution of stimuli that the system saw and over a larger distribution of stimuli. The latter can test whether the system just found a good model of the data it saw or discovered more generalizable theories of the data-generating process. Thus, model recovery analyses are essential to evaluating the success of a system for automated science like this.

In the current paper, we implement an automated theory discovery and experimentation loop for computational cognitive science (Figure \ref{fig:loops}), which we call {\sc auto-psych}. We develop an agent-based scientific discovery framework for creating computational theories and apply it to three classic domains in cognitive science as case studies: subjective randomness, reference games, and the number game. The core of this system is two nested loops: an outer loop that designs experiments and deploys them online (the ``lab in the loop''), and an inner loop that critiques cognitive models and iteratively refines them (Box's loop).

We tested whether {\sc auto-psych} could improve upon leading theories in each domain. To begin, we surveyed the literature in each domain and collected several of the leading cognitive models. We then translated these models to the common framework of probabilistic programs and used them as ``seed'' models for {\sc auto-psych}. We then asked whether {\sc auto-psych} can discover models that fit the human data better than these seed models.

We deployed {\sc auto-psych} on all three domains: agents wrote and critiqued models of cognition and ran chains of experiments with human participants. We found strong theory discovery: across each domain, {\sc auto-psych} finds models that fit the human data better than any of the seed models from the literature. These models can be substantially more complex than previous models, but through ablation studies we show that much of their complexity is necessary to fit the data well. We reflect on the promise of this system as well as potential downsides of using autonomous scientific agents for automating psychological research.

\section*{The {\sc auto-psych} workflow}

Figure~\ref{fig:loops} illustrates the {\sc auto-psych} framework. It consists of two loops: an outer ``lab in the loop'' which designs and deploys human experiments and an inner ``Box's loop'' that critiques and proposes cognitive models. Each model is written as a probabilistic program, a structured statistical model that can be fit to data. The loop maintains a registry of cognitive models, initialized with a set of seed models. On each iteration of the outer loop, {\sc auto-psych} selects the stimuli that maximally distinguish between these models using expected information gain (EIG), then adds those stimuli to an experiment and deploys it on Prolific.

Once the experiment is complete, {\sc auto-psych} pulls the data and passes it to the inner loop for analysis. On each iteration of the inner loop, a critic agent writes test statistics designed to surface significant discrepancies between the human data and the current best-fitting model. A theorist sees the results of these critiques and proposes new cognitive models based on them. At the end of the inner loop, up to twelve models within a threshold value of the best-fitting model are compared. Model fit is evaluated via Expected Log Pointwise Predictive Density (ELPD), a measure of out-of-sample predictive accuracy.

\subsection*{Experimental domains}

We applied {\sc auto-psych} to three classic domains in cognitive psychology: subjective randomness, reference games, and the number game. In each domain, we initialized the loop with a set of seed models inspired by prior literature.

\paragraph*{Subjective randomness}
People have strong intuitions that some sequences of coin flips are more random than others \citep{bar1991perception,ayton2004hot}. For example, {\tt HTTHHHTH}  may seem more random than {\tt HTHTHTHT}, despite the two sequences being equally probable under a fair coin. Previous accounts have explained these judgments in terms of how representative a given sequence is of a prototypical random sequence, its algorithmic complexity, a Bayesian inference about its underlying generating processes, or how likely it is to occur as a subsequence of a long stream of coin flips \citep{kahneman1972subjective,falk1997making,li2008introduction, griffiths2003algorithmic, griffiths2018subjective,hahn2009perceptions}. In our subjective randomness experiments, we gave {\sc auto-psych} a template of a forced-choice paradigm in which participants view two equal-length sequences of coin flips and choose which one appears more random. This setting is well suited for automated model discovery because the stimulus space is discrete and tractable to enumerate, the experimental paradigm is straightforward, and computational models can be used to predict the probability of choosing either sequence.

\paragraph*{Reference games} People make inferences about the meanings of utterances that go well beyond their literal meanings \citep{grice1975logic,clark1996using}. Cognitive scientists have studied this kind of reasoning using reference games, where participants are presented with an utterance and a set of possible referents (e.g., shapes with different colors). Their goal is to guess which of the items the utterance refers to. For example, a participant might see a blue square, a green square, and a green circle and the utterance ``green.'' The participant would then need to choose which referent the utterance refers to. Participants usually choose the green square, since a speaker would presumably have said ``circle'' if they meant to refer to the green circle. The Rational Speech Acts (RSA) framework predicts the choices participants make in these games through models of recursive mental reasoning \citep{frank2012predicting,goodman2016pragmatic,degen2023rational}. Many variants of this model have been developed to capture subtle linguistic phenomena including the non-literal use of number words \citep{kao2014nonliteral}, politeness \citep{yoon2020polite}, and generic language \citep{tessler2026learning}. The {\tt memo} programming language was recently designed to efficiently compute the kind of recursive mental reasoning that RSA models use \citep{chandra2025domain}. For our purposes, this domain consists of up to four items with different features that are present or absent (e.g. a scarf, glasses, and a hat) and an utterance that names one of the features (e.g. ``glasses''). Participants choose the item that they think the speaker was most likely referring to.

\paragraph*{Number game} This is a classic task used to study how people generalize concepts from a small number of examples \citep{tenenbaum1999bayesian, tenenbaum1999rules, tenenbaum2001generalization}. Participants are shown a set of numbers belonging to an unknown category, such as $\{8,16,32\}$. They are then asked to judge whether a new number, such as $24$ or $64$, belongs to the same category. The original set of numbers is consistent with multiple hypotheses, such as ``even numbers'' or ``powers of two,'' which make different predictions about which of the unseen numbers belong to the same category. Bayesian models of concept learning explain these judgments as inferences over possible concepts, which combine prior expectations about these categories with the evidence provided by the observed examples \citep{tenenbaum1999bayesian, tenenbaum1999rules, tenenbaum2001generalization}. Our study uses an existing, large-scale dataset collected by Bigelow and Piantadosi \citeyearpar{bigelow2016large}, in which participants provided binary (yes/no) judgments for whether a new number belonged to the same category.

\begin{figure}
    \centering    
    \includegraphics[width=\textwidth]{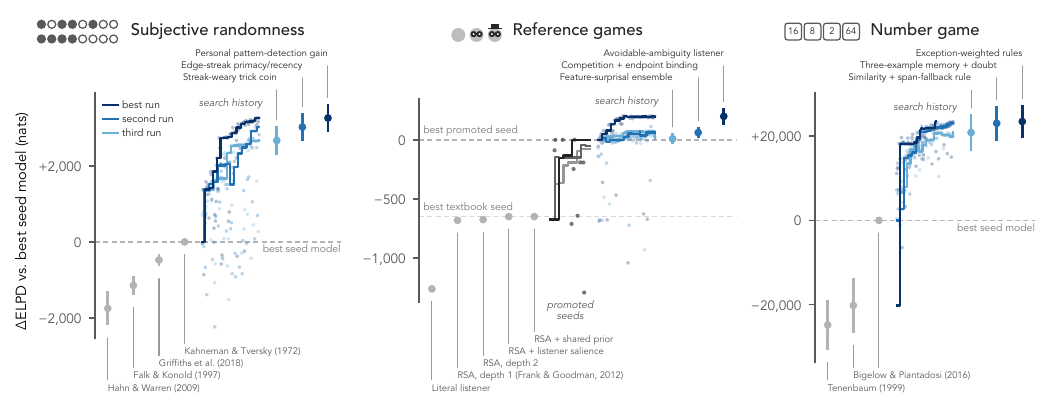}
    \caption{Expected log pointwise predictive density (ELPD) of seed models (grey) and models written by {\sc auto-psych} in each domain. All models were fit to the pooled set of data collected in each domain. Subjective randomness models were also evaluated on that data. Reference games models were evaluated on data from the preregistered follow-up experiment. Number game models were evaluated on all responses to the stimuli that were not tested in any run. The reference games panel first shows the seed promotion on existing data in grey, then the live runs in blue. Error bars show two stimulus-clustered standard errors.}
    \label{fig:elpd}
\end{figure}

\section*{Results}

Before running {\sc auto-psych} on our domains, we validated its ability to recover ground-truth models by holding one seed model out of the starting set at a time and using it to generate ground-truth responses. In a series of model recovery simulations across all three domains, {\sc auto-psych} reliably discovered models close to the ground truth, as measured by root mean squared error (RMSE) over all possible stimuli. The agent-written models consistently showed lower RMSE than the best other seed model fit to the same data (SI Appendix~\ref{app:recovery}).

In the subjective randomness domain, we conducted a further validation of model recovery by creating ``alien'' ground truth models that are very different from anything in the existing literature. For example, one of these models judges sequences of coin flips as more random if they have more heads (vs. tails). We found even stronger recovery for these models than for the held-out seed models (SI Appendix~\ref{app:recovery}). 

We next ran {\sc auto-psych} on our three domains, collecting new data from human participants in both subjective randomness and reference games. We conducted three runs in each domain; each run consisted of three iterations of the outer loop. In addition to the main model comparisons, we ran comparisons using Akaike information criterion, Bayesian information criterion, and Bayes factor, which can be found in SI Appendix~\ref{app:aic-bic}.

\begin{figure}[tbp]
    \centering
\includegraphics[width=\linewidth]{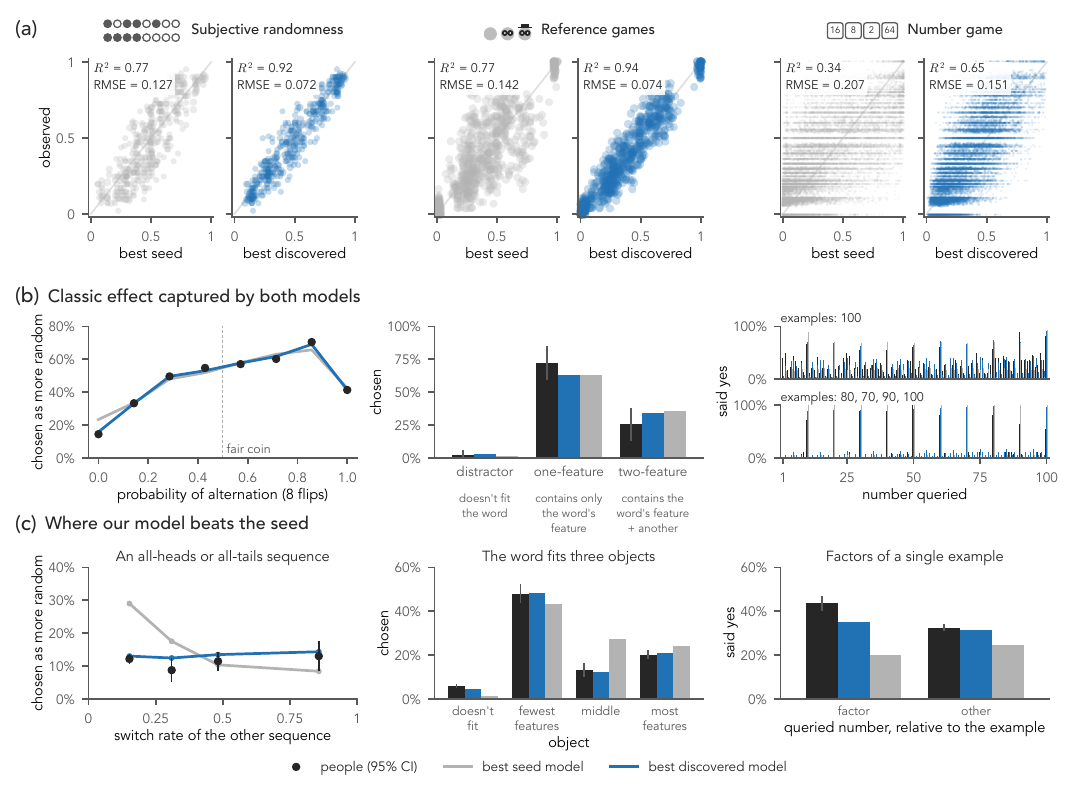}
    \caption{Comparison of the behavior of seed and discovered models. \textbf{(a)}: stimulus-level relationships between the best seed (grey) and winning model (blue) in each domain. \textbf{(b)}: a classic effect captured by both the seed and discovered models: sequences with more alternations (but not strictly alternating) are judged as more random in subjective randomness, participants pick the referent whose only feature matches the utterance (e.g. glasses) rather than the referent without a matching feature (no glasses) or an additional feature (hat and glasses), both models match the general shape of the human generalization profile for one and four examples in number game. \textbf{(c)}: A new effect captured by the winning model but not the best seed: participants' probability of picking a sequence of all heads/tails is relatively flat over different alternation rates for the other sequence, participants favor choosing candidates with the fewest features over those with intermediate numbers of features, and in the number game participants were more likely to make ``yes'' judgments when the example set was only one number and the query number was a factor of the example number. These new effects were found via an exploratory post-hoc search over features of the stimuli. Reference games uses the best textbook seed. Error bars show 95\% confidence intervals.}
    \label{fig:stimulus}
\end{figure}

\subsection*{Subjective randomness}

We started our subjective randomness runs with four seed models: 1) an ``encoding difficulty'' model that uses the difficulty of encoding a sequence in a compressed representation to represent randomness judgments \citep{falk1997making}, 2) a ``regularity detection'' model that compares the probability that the sequence was generated by a fair coin against the probability it was generated by a more regular computational process \citep{griffiths2003algorithmic,griffiths2018subjective}, 3) a ``finite experience'' model inspired by \citet{hahn2009perceptions} that judges a sequence based on how likely it is to occur as a subsequence of a longer sequence of fair coin flips, and 4) a ``local representativeness'' model that judges sequences' randomness based on how well their heads/tails imbalance and alternation rate resemble a mental prototype, as well as the representativeness of each local group of a few outcomes \citep{kahneman1972subjective,reimers2018perceptions}. These models are designed to be faithful to the accounts developed in the original papers, though the models are not necessarily identical since we adapted them to the PyMC framework and some papers did not state their theories as Bayesian models. We then ran three independent replicates of the {\sc auto-psych} workflow. Further details of the seed models can be found in SI Appendix~\ref{app:seeds}.

Across all nine experiments, 359 unique participants completed 361 short experiments, giving us 23,104 trials. 416 of 576 stimulus pairs (72\%) were unique, indicating that the same stimulus pairs were largely not being used across experiments or across runs.\footnote{Due to a bug in computing ELPDs, 6 of the 8 carried-over models were dropped at the beginning of the second experiment in run 2.} Most chosen displays (453 of 576) showed sequences with the maximum number of coin flips. Participants showed no meaningful side bias (mean=0.50, 95\% CI [0.495, 0.511]) and no meaningful heads bias (0.50 [0.49, 0.51]) but they did rate sequences with more alternations than chance as more random (0.66 [0.64, 0.68]), which is consistent with prior studies \citep{kahneman1972subjective}.

Each of the three replicates discovered a model that explained the human data much better than the seed models according to ELPD. The fits of the best model and the seed models to the pooled dataset from all three replicates are shown in Figure~\ref{fig:elpd} (left). 

The best-performing model on the pooled dataset across all replicates was called ``Personal pattern detection gain.'' It compares a sequence's alternation rate to a fitted individual ideal alternation rate, along with a penalty for imbalances between the number of heads and tails and an adjustment for the length of the longest streak of the same outcome. It captures several trends in the human data that the seed models miss, including individual variability in how much a participant penalizes the length of the longest streak. It also captures a preference for a variety of run lengths. This model draws upon ideas from \citet{kahneman1972subjective}, such as the fact that people think maximally random sequences have alternation rates higher than 0.5 and they view imbalances as evidence against randomness. However, it also draws upon \citet{griffiths2018subjective} in that some of its randomness penalties are indicators that the sequence was generated by a structured process rather than a sequence of fair coin flips. The final model is much more complex than the seed models, having 17 group-level parameters and 6 individual-level parameters per participant. In contrast, the most complex seed model has seven group-level parameters and no individual-level parameters.

We analyzed the winning model's fit at the level of specific stimulus pairs and found that the winning model made predictions that correlated with human judgments more strongly than the best seed model, and captured classic effects from prior psychology literature. In an exploratory analysis, we looked for stimulus-level patterns in human behavior that the winning model captures but the best seed does not. We found that participants' responses to stimulus pairs that included one sequence of all heads or all tails were consistently low regardless of what the other sequence was, while the local representativeness model predicts participants to be closer to indifferent over a sequence of all heads or all tails and another sequence with a low alternation rate. See Figure~\ref{fig:stimulus}c for this analysis.

\subsection*{Reference games}

In the reference games domain, we seeded {\sc auto-psych} with five models from ProbLang, an online textbook on probabilistic models of language understanding, written in memo \citep{scontras2018probabilistic,hawkins2025problangv2}. This domain has a wealth of open data available, so we first made use of it by applying the inner loop of {\sc auto-psych} to a combined dataset of 50,587 trials from five papers in the literature with open data available \citep{frank2016rational,sikos2021reevaluating,mayn2022individual,mayn2023high,mayn2026sources}. We grouped all of the models that were admitted over the course of these runs into clusters based on what predictions they make about how people would behave, then took the best-fitting model within each cluster as a seed model. This left us with 10 models. We also added the canonical depth 1 and depth 2 RSA listener models from the ProbLang textbook \citep{goodman2016pragmatic}.

We then ran three separate runs of {\sc auto-psych}, each with five or six seed models chosen in order to give each start a diverse set (See SI Appendix~\ref{app:more-human-details} for more details). This resulted in a total of nine short experiments (1,659 participants, 21,567 trials included after catch trial exclusions). Again, {\sc auto-psych} produced models that out-performed the seeds (Figure~\ref{fig:elpd}).

We then ran a large-scale preregistered follow-up experiment comparing the best discovered model from each loop to the seed models using parameters fit on the full previous dataset (including both the nine experimental runs and the data from prior literature).\footnote{The preregistration for this experiment can be found at \url{https://osf.io/s48jv}.} Stimuli were chosen half according to EIG over the seed and winning models and half randomly. For this experiment, we included data from 1,108 participants and 14,404 trials. The main preregistered analysis was a set of three one-sided Holm-corrected z-tests comparing the best discovered model in each run with the best-fitting promoted seed model. Standard errors were clustered by display. The discovered models from run 1 ($z=5.64, p < 0.0001$) and run 3 ($z=2.66, p = 0.008$) significantly outperformed the best promoted seed, but the discovered model from run 2 did not ($z = 0.34, p = 0.37$). We also preregistered separate comparisons on the random and EIG halves of the dataset, reported in SI Appendix~\ref{app:prereg-half-comparison}. The best-fitting model here also provided a better stimulus-level fit to the data from the preregistration than the best textbook seed model did (Figure~\ref{fig:stimulus}). 

The best-fitting discovered model was the ``Avoidable-ambiguity listener'' model from run 1. This model posits an L2, a listener who reasons about a pragmatic speaker, who in turn reasons about a depth-1 pragmatic listener. The listener has a saliency prior over referents which favors objects with more features. If the utterance names a feature that applies to three or four stimuli, the listener favors the item with either the fewest or the most features rather than the item with an intermediate number of features (see Figure~\ref{fig:stimulus}c). This model has 17 parameters, compared to a maximum of 7 for the promoted seed models.

\subsection*{Number game}

While the agents in the other two domains ran live experiments, we used the number game domain to test whether {\sc auto-psych} can effectively discover new insights about existing large datasets. \citet{bigelow2016large} gathered a large dataset of 272,700 human responses to different number game queries. Rather than running new experiments, in these runs {\sc auto-psych} chooses which data points to reveal from the larger set. Specifically, in each outer loop round, the loop chooses 280 pairs consisting of an example set and a query number (e.g., the set $\{16, 32, 2, 8\}$ and the query number $64$). It then samples 9 previously-unseen human responses from each pair, without replacement. This process mirrored stimuli selection and data collection in the standard {\sc auto-psych} loop, but only using existing data.

We seeded the number game runs with three models, the Bayesian model developed by \citet{tenenbaum1999rules}, which predicts by marginalizing over a space consisting of rule-based (e.g., "powers of $2$") and interval-based (e.g., "the interval $20-30$") hypotheses, and uses a likelihood function based on the "size principle", which states that smaller hypotheses (e.g., "multiples of $10$") are more likely than larger ones (e.g., "multiples of $2$") when both are consistent with the data. We also included two models developed by \citet{bigelow2016inferring}. The first is a model similar to that proposed by \citet{tenenbaum1999rules}, but with additional rules (e.g., "contains digit $n$") and a uniform prior over intervals in place of Tenenbaum's Erlang prior. The second has an identical hypothesis space, but rather than an equal prior on all rule-based hypotheses, it includes a fitted prior parameter for each rule.

Again, {\sc auto-psych} discovered models that significantly out-performed the best-fitting seed model when fit to the pooled data in terms of ELPD on the remaining dataset (Figure~\ref{fig:elpd}). It also fit human behavior better at the stimulus level (Figure~\ref{fig:stimulus}). The best-fitting model on the unseen data, called ``exception-weighted rules,'' came from run 3. The ``three-example memory + doubt'' model from run 2 came quite close and fit better on the data seen during the runs. We analyze ``exception-weighted rules'' here. This model is a variant of a Bayesian learner like the seed models, but it makes several changes to the seeds: it treats each example as weaker evidence than an ideal Bayesian observer does. Some concepts, like ``numbers above 50'' receive higher prior weight than the seed models give them. The model captures patterns in the data that the seeds do not. For instance, when people see a single example they are more likely to judge a query number as part of the set if it is a factor of the example number than the best seed model predicts. The exception-weighted rules model captured this trend better, though there is still some gap between the human average and the model's prediction. Figure~\ref{fig:stimulus} shows this trend.

\subsection*{Ablations of winning models}

\begin{figure}
    \centering
    \includegraphics[width=0.9\linewidth]{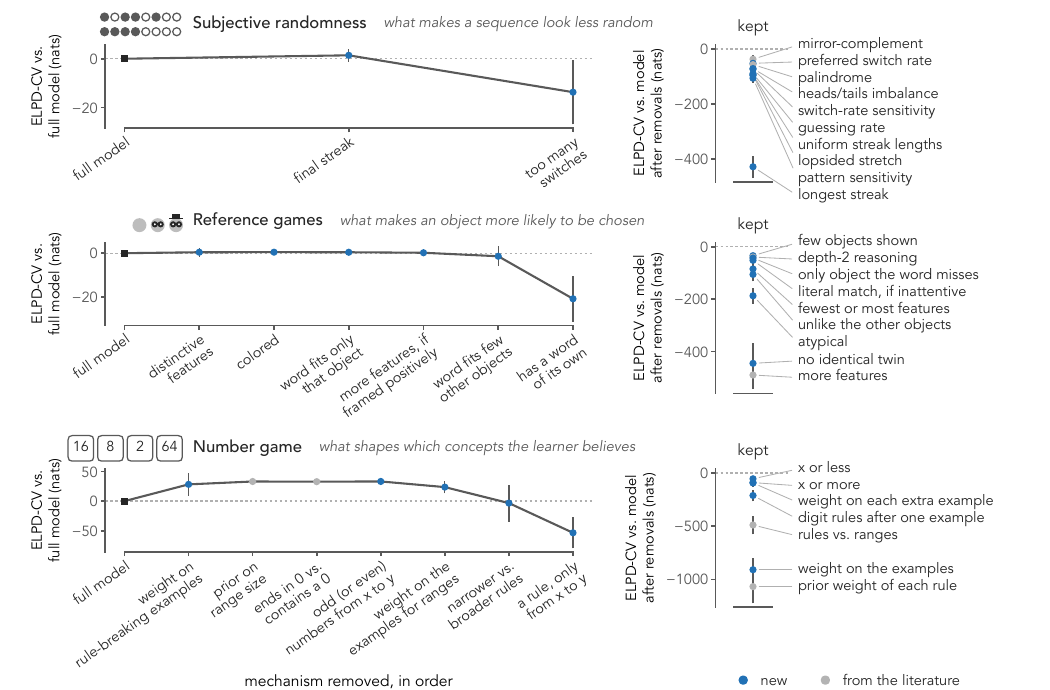}
    \caption{Ablations of winning models in each domain. Each point shows the best-performing model after each round of backward elimination. Error bars show one stimulus-clustered standard error of the difference between each model and the best model from the last round. The right panel shows the mechanisms that remained in the model after ablations. Points are labeled with the name of the feature that was removed in that round.}
    \label{fig:ablations}
\end{figure}

While the models that {\sc auto-psych} discovered achieved strong fits to the human data, the models in the subjective randomness and reference games domains have notably more parameters than the seed models. How much of this complexity is necessary to capture patterns in the human data and how much of it could be removed? To answer this, we conducted ablation studies via backward elimination. In our ablations, an AI agent rewrote the winning model in a format where each mechanism could easily be turned on or off. After validating that these rewritten models make the same predictions as the original models, we performed backward elimination on the mechanisms. We tried every possible variant of the model that removed one mechanism (including removing individual-level variability in parameters). We then simplified the model to be the best-performing of these reduced models in terms of ELPD computed via 5-fold cross-validation on the full dataset. We repeated this process, removing one mechanism at a time, until the best-performing ablated model performed two standard errors worse than the best model from the previous round.

Figure~\ref{fig:ablations} shows the results of this analysis. For all of the winning models, it is possible to remove a few mechanisms without a substantial cost to predictive accuracy. For example, the first five mechanisms removed from the winning reference games model produced almost no difference in fit. These mechanisms included a salience term based on whether a stimulus was in color or monochromatic which did not matter because none of the stimuli were monochromatic. The first ablation of the winning number game model actually improved fit slightly. After ablating a few mechanisms, however, model fit starts to decline quickly, and the ablated models still have more parameters than any seed model in the subjective randomness and reference games domains. Thus, while some of the winning models' complexity was unnecessary, much of the increase in complexity from the seed models proved important to fitting the data well. Future work should consider how to enforce parsimony constraints even more stringently during the theory discovery process.

\section*{Discussion}

In this paper, we introduced {\sc auto-psych}, a framework for automatically generating, testing, and critiquing computational models of cognition. The core design feature of this framework is the presence of two nested loops: an inner loop that proposes and critiques models and an outer loop that designs and runs experiments with human participants to differentiate models. We applied {\sc auto-psych} to three classic domains in cognitive science and found that it could recover ground-truth models informed by the literature as well as psychologically alien models. We used the framework to implement and run experiments with real human participants, finding that it discovered cognitive models that fit human behavior significantly better than the initial seed models. Because these seed models are inspired by influential theories of the target domain, this result indicates strong discovery ability. Our framework builds upon prior methods for the automated discovery of cognitive models by adding live experimentation. At a larger scale, a process like this might be able to explore vast spaces of experimental designs autonomously and find better models of human cognition. 

The agent-discovered models included novel revisions to the models which fit the human data much better than the seed models, including to data that were never seen in their discovery loops. Two of these winning models were much more complex than the models that we seeded the loop with. Ablation studies indicated that several mechanisms could be removed without a significant cost to fit, but some of the added complexity was important to help the models achieve strong fits to the data. Future work should focus on ways of analyzing a range of agent-discovered models to extract more insights about human cognition.

\subsection*{Limitations}

Like other automated scientific discovery systems \citep{swanson2025virtual,lu2026towards}, our work provides a proof of concept for feasibility using three case studies. These are all classic paradigms in cognitive science with a rich set of theories in the literature. Nevertheless, the stimulus spaces are highly restricted, making it simpler to create experiments and choose stimuli than in many commonly used paradigms. The system we designed is quite general, and LLM-based coding agents are improving rapidly. Therefore, we expect that our work could be applied to other paradigms, but scaling for any given paradigm will be controlled by the unevenness of the theoretical space as well as the richness of the stimulus and experimental design options.

Our work also shares the limitations of much prior work in computational cognitive science. The field's focus has primarily been on providing parsimonious high-level descriptions of human cognition. Most classic investigations focused on schematic stimuli, simplified theories, participant averages rather than individual variation, and measurements of convenience samples from ``WEIRD'' populations \citep{kroupin2025beyond}. We adopted this general -- intrinsically limited but still powerful -- approach here, recognizing that modern work attempts to circumvent these limitations, including through the use of naturalistic stimuli, richer theoretical vocabularies, and models of individual participants \citep{carvalho2025naturalistic,peterson2021using,lee2005modeling,tauber2017bayesian, fan2026generative}.

An additional limitation of our approach strikes at the core of psychological science: those models discovered by {\sc auto-psych} may fit data better than human-created models and may even be measurably more parsimonious (e.g., shorter). Yet these models may still fail to provide satisfying explanations for human psychologists. Whether deeper explanatory virtues can or should be available to automated systems remains to be seen.

Our experiments selected stimuli using Expected Information Gain with respect to a set of models under consideration. However, recent simulation work has shown that randomly choosing experimental designs can lead to better estimates of a ground-truth quantity than choosing designs that maximally discriminate between theories, at least under some circumstances \citep{dubova2026against}. We expect our system could be an important tool for adjudicating questions like these in future work. 

\subsection*{Conclusion: The promise (and perils) of automated scientific discovery}

A better understanding of human cognition could lead to important advances in education, clinical treatment, and human-computer interaction. Scientific automation has serious potential costs, however. One of these is the threat of increasing homogeneity in scientific theories \citep{khosrowi2026automating,hao2026artificial}. If the only theories scientists consider are those proposed by the same set of theorist agents, they might easily overlook good alternatives that are low probability for those agents. Additionally, if AI-driven science led to a flood of publications that human scientists had to review, it could further overwhelm an already stretched peer-review infrastructure \citep{messeri2026uncritical}. Finally, automating science removes training opportunities for scientists, potentially leading to ``deskilling'' that might decrease the set of individuals qualified to judge the outputs of automated discovery systems \citep{lenharo2026ai,shen2026aiimpactsskillformation}. 

Mitigating these issues will require thoughtful integration of discovery systems with existing scientific infrastructure. Rather than replacing scientists, small-scale discovery systems such as {\sc auto-psych} might be a starting point for scientists to compare theories across phenomena, seek integrative models, or iterate before exploring more costly physiological or neural measurements. Software innovations -- such as those we use here for creating probabilistic programs or launching web experiments -- have routinely raised the level of abstraction at which scientists can work. We hope that automated discovery tools can provide the next layer of abstraction to accelerate progress in understanding the mind.

\section*{Materials and Methods}
Our framework uses LLM-based agents in coding harnesses to instantiate two loops: an inner loop that conjectures, fits, and critiques models and an outer loop that designs and runs experiments. For the subjective randomness and number game domains, we used Claude Opus 5.5 with the Claude Code harness \citep{anthropic2026claudeopus55}. For the reference games domain, we used Gemini 3.8 Flash with the OpenCode harness. The {\sc auto-psych} loop has two agents involved in developing computational models: a theorist agent that conjectures new models and a critic agent that looks for discrepancies between the human data and the current best-fitting model. Optionally, an implementation agent implements online experiments, though experiment implementation could also be done programmatically. Code and data are available on \href{https://github.com/mcfrank/auto-psych}{GitHub}. Further methodological details can be found in SI Appendix~\ref{app:more-details}.

\subsection*{Theory framework}

In our framework, a cognitive model is instantiated as a probabilistic program: snippets of Python code written in a library that enables easy probabilistic inference. The subjective randomness and number game experiments use PyMC, a general-purpose probabilistic programming language \citep{abril2023pymc}. The reference games use memo, a probabilistic programming language based on Python and optimized specifically for recursive mental reasoning \citep{chandra2025domain}. Each program defines a distribution over participant choices in a given trial as a function of the stimuli presented in that trial. These models can also write functions that preprocess the sequences to compute useful features. In the subjective randomness domain models can be hierarchical, with person-level parameters in addition to group-level parameters. We place weak priors over these free parameters and infer distributions by fitting the models to observed data. 

\subsection*{Modeling loop}

The {\sc auto-psych} workflow consists of an outer loop that chooses stimuli to optimally distinguish between models and implements experiments and an inner loop that critiques and improves upon models of the participant data. {\sc auto-psych} is initialized with a set of \textit{seed models}. We then scored stimuli using Expected Information Gain with respect to the models under consideration and selected them greedily to maximize their joint EIG~\citep{ouyang2016practicaloptimalexperimentdesign,foster2019variationalbayesianoptimalexperimental,gandhi2025boxing}. We compute EIG using a uniform prior over models in the registry on each step. In the reference games domain, we used a prior that was a mixture of a uniform distribution over models in the registry and a uniform distribution over the baseline models we compared against.

After choosing stimuli, an implementation agent writes an experiment using jsPsych. We gave these agents strict templates to follow and automatically added our consent form to the experiment code.\footnote{In the subjective randomness experiment, there was an error in the template which meant that trial order was not randomized. Trials were presented in the order of EIG.} The pipeline deploys the experiment to Firebase and recruits participants using the Prolific API. It then polls the Prolific study to monitor when it is completed, pulls the data, and continues the loop. All experiments were approved under Stanford IRB protocol~\#20009 and participants completed a consent form before the experiments. More details about the human experiments can be found in SI Appendix~\ref{app:more-human-details}.

After collecting data, the workflow proceeds to the inner loop. This is an implementation of Box's loop: a framework for iteratively improving statistical models that alternates between critiquing the current best model and positing new models to address those critiques. Each iteration of this loop starts with a critic agent adapted from the CriticAL framework~\citep{li2024criticalcriticautomationlanguage}. The critic conjectures eight test statistics designed to surface differences between the posterior of the best-fitting model and the human data. The theorist then sees the results of any test statistics that surfaced a significant difference at the $p=0.05$ threshold and proposes new models to improve fit to the data. The theorist proposes models in a manner inspired by the ``hypothesis search'' framework: the agent first writes a theory in natural language, then translates it to code \citep{wang2024hypothesis}. This loop repeats up to five times, proposing six models per iteration. Different theorist agents receive different goals at each iteration. For instance, some agents are prompted to explore a new idea not tried so far, while others are prompted to refine the incumbent model. Models must pass a series of checks, including producing finite log-likelihoods, Markov chain Monte Carlo convergence, and not being too close to any existing model in the registry. At the end of all iterations of the inner loop, all models whose ELPD is lower than the best model by at least two stimulus-clustered standard errors are removed from the registry.\footnote{In subjective randomness, we raised this threshold to four standard errors after the first experiment of run 1 would have pruned all but one model.}
If more than eight models are within this range, only the top eight are kept. These remaining models are added to the model registry. Further details about how this loop works can be found in SI Appendix~\ref{app:more-details}.

\backmatter{Data, Materials, and Software Availability}Code and data are available publicly at \url{https://github.com/mcfrank/auto-psych}.

\backmatter{Acknowledgments}Thanks to Google, Inc. for gift funds and Gemini compute credits that were used to support this work. This research is partially supported by the DARPA expMath program through the DARPA CMO contract number HR0011262E020. We would like to thank Dr. Patrick Shafto, expMath Program Manager, for useful technical discussions. Some of the computing for this project was performed on the Sherlock cluster. We would like to thank Stanford University and Stanford Research Computing for providing computational resources and support that contributed to these research results. We also thank Eric Bigelow and members of the Stanford Computation and Cognition lab for helpful discussions. AI agents were used to write code, check the reproducibility of results, and provide feedback on drafts of this paper. Cursor Agent, Claude Opus 4.8, 5, and 5.5, Claude Fable 5.1, and GPT 5.6 Sol were used to write code. Claude Opus 4.8, 5, and 5.5, Fable 5.1 and 5.2, and GPT 6 Astra were used to review the code and paper. All text was written by humans, with the exception of values in tables, prompts, and pseudocode.

\backmatter{Author contributions}B.P.: Conceptualization, Methodology, Software, Validation, Investigation, Writing - Original Draft, Writing - Editing and Review, Visualization; K.M.: Conceptualization, Methodology, Software, Investigation, Writing - Original Draft; D.W.: Conceptualization, Methodology, Software, Investigation, Writing - Original Draft; L.N.: Conceptualization, Methodology, Writing - Original Draft; M.Y.L.: Conceptualization, Methodology, Writing - Editing and Review; N.D.G.: Conceptualization, Methodology, Software, Writing - Editing and Review, Supervision; M.C.F.: Conceptualization, Methodology, Software, Validation, Investigation, Resources, Data Curation, Writing - Original Draft, Writing - Editing and Review, Visualization, Supervision, Project Administration, Funding Acquisition

\backmatter{Competing interests}The authors declare no competing interests.

\bibliographystyle{plainnat-keepcase}
\bibliography{auto-psych}

\clearpage
\section*{Supporting Information Text}

\appendix
\setcounter{figure}{0}
\setcounter{table}{0}
\renewcommand{\thefigure}{S\arabic{figure}}
\renewcommand{\thetable}{S\arabic{table}}
\renewcommand{\theHfigure}{S\arabic{figure}}
\renewcommand{\theHtable}{S\arabic{table}}

\section{Further methodological details}
\label{app:more-details}

This section provides further methodological details on both loops of the {\sc auto-psych} workflow. Algorithm~\ref{alg:outer} shows pseudocode for the outer loop and Algorithm~\ref{alg:inner} shows pseudocode for the inner loop. Many of the specific implementation details varied across different domains. The details of these differences are shown in Table~\ref{tab:domain-methods}.

\subsection{Outer loop}

The outer loop iteratively designs experiments to collect data that would be maximally informative in adjudicating between hypotheses~\citep{gandhi2025boxing}. We begin the first experiment cycle with a set of ``seed'' models (per above). Then, given a set of models $\mathcal{M}$ and data $\mathcal{D}$, the goal of the outer loop is to run new experiments $N$ that optimally distinguish between the current set of models under consideration.

The first iteration of the loop begins by selecting stimuli to optimally distinguish between seed models. Subsequent iterations select stimuli to distinguish between models in the registry. We use a uniform prior over the models in the registry when computing EIGs. The goal of stimulus selection is to maximize the joint EIG of the set of stimuli. We solve this problem approximately by greedily selecting the stimulus that maximizes the joint EIG conditioned on simulated responses to the previous stimuli added to the set. We compute the EIG of a single stimulus based on the distribution over possible participant responses. We do this using draws from each model's predictive distribution -- the prior predictive in the first iteration and the posterior predictive in subsequent iterations. The posterior is thinned to 200 draws to compute predictions. For models with person-level parameters, we drew new sets of participant parameters from the fitted distribution and averaged predictions over those participants. The objective we maximize is
\begin{equation}
    \mathrm{EIG}(S) = H(M) - \mathbb{E}_{\mathbf{k}_S \sim P(\mathbf{k}_S)}\big[ H(M \mid \mathbf{k}_S) \big]
\end{equation} 
\noindent where $H(M)$ is the entropy over models prior to a particular stimulus and $\mathbf{k}_s$ is the vector of counts of the participants' responses when presented with stimulus $s$. $H(M \mid \mathbf{k}_s)$ is the entropy over models, conditioned on those responses. 

We selected greedily until the estimated EIG of the best next stimulus is within two Monte Carlo standard errors of 0. After this, stimuli were selected based on their single-response EIG. We recomputed EIGs for each stimulus lazily: we re-scored only the top 512 stimuli each round to save compute, but we re-scored the full set of stimuli after every 16 stimuli were chosen.

Once stimuli are chosen, the outer loop deploys the experiment to Firebase and recruits participants on Prolific. Once the study is complete, the data is passed to the inner loop for analysis.

\begin{algorithm}[tbp]
\caption{Outer loop: model-guided experiment design and data collection.}
\label{alg:outer}
\begin{algorithmic}[1]
\Statex \textbf{Require:} seed models $\mathcal{M}_0$; optional existing data $\mathcal{D}_0$; stimulus pool $\mathcal{C}$; optional baseline models $\mathcal{B}$
\Statex \textbf{Parameters:} experiments $N$; stimuli per experiment $n_S$; responses per stimulus $n_R$; design-prior weight on carried models $w$ ($1$ for SR and NG; $0.5$ for RG); models carried between experiments $n_{\max}$
\Statex
\Function{GreedyEIG}{$\mathcal{C}, P, n_S$}
  \State $\mathcal{S} \gets \emptyset$
  \Repeat \Comment{joint EIG, by Monte Carlo over $m \sim P$, $\theta \sim p(\theta \mid m, \mathcal{D})$, $n_R$ responses per stimulus}
    \State $g(c) \gets \mathrm{EIG}(\mathcal{S} \cup \{c\}) - \mathrm{EIG}(\mathcal{S})$ for each $c \in \mathcal{C} \setminus \mathcal{S}$
    \If{$\max_c g(c) \le 2\,\mathrm{SE}_{\mathrm{MC}}$} \textbf{break} \EndIf
    \State $\mathcal{S} \gets \mathcal{S} \cup \{\arg\max_c g(c)\}$
  \Until{$|\mathcal{S}| = n_S$}
  \While{$|\mathcal{S}| < n_S$} \Comment{fill by single-response EIG, given $\mathcal{S}$}
    \State $\mathcal{S} \gets \mathcal{S} \cup \{\arg\max_{c} \mathrm{EIG}_1(c \mid \mathcal{S})\}$
  \EndWhile
  \State \Return $\mathcal{S}$ \Comment{RG: each pick must keep the display-type quotas reachable}
\EndFunction
\Statex
\State $\mathcal{D} \gets \mathcal{D}_0$;\quad $\mathcal{M} \gets \mathcal{M}_0$
\For{$n = 1$ \textbf{to} $N$}
  \If{$n > 1$}
    \State $\mathcal{M} \gets$ the best $n_{\max}$ models in $\mathcal{M}_{\mathrm{live}}$, merging near-duplicates \Comment{SR and NG skip this step}
  \EndIf
  \State $\mathcal{B}' \gets \mathcal{B}$ without duplicates of models in $\mathcal{M}$ or of each other
  \State $P \gets w \cdot \mathrm{Uniform}(\mathcal{M}) + (1-w) \cdot \mathrm{Uniform}(\mathcal{B}')$
  \State draw 200 parameter samples for each model in $\mathcal{M} \cup \mathcal{B}'$ given $\mathcal{D}$ \Comment{prior if $\mathcal{D} = \emptyset$}
  \State $\mathcal{S} \gets \textproc{GreedyEIG}(\mathcal{C}, P, n_S)$
  \State $\mathcal{R}_n \gets \textproc{Collect}(\mathcal{S})$
  \State $\mathcal{D} \gets \mathcal{D} \cup \mathcal{R}_n$
  \State $\mathcal{M}_{\mathrm{live}}, m^\dagger \gets \textproc{InnerLoop}(\mathcal{D}, \mathcal{M})$
\EndFor
\State \Return $\mathcal{M}_{\mathrm{live}}$, $m^\dagger$
\end{algorithmic}
\end{algorithm}

\begin{algorithm}[tbp]
\caption{Inner loop: conjecture, fit, and critique cognitive models.}
\label{alg:inner}
\begin{algorithmic}[1]
\Statex \textbf{Require:} pooled data $\mathcal{D}$; carried models $\mathcal{M}$
\Statex \textbf{Parameters:} rounds $R$; candidates per round $K$; critique statistics $J$; significance level $\alpha$; posterior-predictive replicates $B$; pruning threshold $k$; live-set cap $L$
\Statex
\Function{Score}{$\mathcal{M}, \mathcal{D}$}
  \For{$m \in \mathcal{M}$}
    \State fit $m$ to $\mathcal{D}$ by Markov chain Monte Carlo
    \State $s(m) \gets$ estimated ELPD
    \State $\mathit{trusted}(m) \gets$ $m$ converged and $s(m)$ is reliable
  \EndFor
  \State \Return $s$, $\mathit{trusted}$
\EndFunction
\Statex
\State $\mathcal{M} \gets \{m \in \mathcal{M} : m \text{ fits } \mathcal{D} \text{ with finite } s(m)\}$
\State $s, \mathit{trusted} \gets \textproc{Score}(\mathcal{M}, \mathcal{D})$
\For{$t = 1$ \textbf{to} $R$}
  \State $m^\dagger \gets \arg\max_{m \,:\, \mathit{trusted}(m)} s(m)$
  \State \textproc{Critic} writes test statistics $T_1, \dots, T_J$ for $m^\dagger$ \Comment{RG: computed on live trials only}
  \For{$j = 1$ \textbf{to} $J$}
    \State $t_j^{\text{obs}} \gets T_j(\mathcal{D})$;\quad
      $\{t_j^{(b)}\}_{b=1}^{B} \gets T_j(\tilde{\mathcal{D}}^{(b)})$,\ \
      $\tilde{\mathcal{D}}^{(b)} \sim p(\cdot \mid m^\dagger, \mathcal{D})$
    \State $p_j \gets$ two-sided empirical $p$ of $t_j^{\text{obs}}$ against $\{t_j^{(b)}\}$
  \EndFor
  \State $\mathcal{F} \gets \{\,T_j : p_j \le \alpha\,\}$
  \For{$i = 1$ \textbf{to} $K$ \textbf{in parallel}} \Comment{3 explore, 2 refine $m^\dagger$, 1 refines a model of its choice}
    \State $h_i \gets \textproc{Theorist}$ states a cognitive theory, given $\mathcal{M}$, $s$, and $\mathcal{F}$
    \State $m_i \gets$ probabilistic program implementing $h_i$
  \EndFor
  \State add to $\mathcal{M}$, in slot order, each $m_i$ that passes the checks \Comment{runs, converges, finite likelihood on $\mathcal{D}$, novel}
  \State revise each failed $m_i$ once, in parallel, given the reason it failed
  \State add to $\mathcal{M}$, in slot order, each revision that passes the checks
  \State $s, \mathit{trusted} \gets \textproc{Score}(\mathcal{M}, \mathcal{D})$
  \If{the best trusted model has not changed for 2 rounds} \textbf{break} \Comment{RG only}
  \EndIf
\EndFor
\State $m^\dagger \gets \arg\max_{m \,:\, \mathit{trusted}(m)} s(m)$
\State remove from $\mathcal{M}$ every model more than 4 SE behind the best on published data \Comment{RG only}
\State remove from $\mathcal{M}$ every trusted $m$ with $s(m^\dagger) - s(m) > k \cdot \mathrm{SE}\big(s(m^\dagger) - s(m)\big)$ \Comment{SE clustered by stimulus/display}
\While{$|\mathcal{M}| > L$}
  \State remove an untrusted model if any, else the lowest-scoring model other than $m^\dagger$
\EndWhile
\State \Return $\mathcal{M}$, $m^\dagger$
\end{algorithmic}
\end{algorithm}

\subsection{Inner loop: Theory proposal and critique}

We now define the inner loop, in which theories are proposed and critiqued on existing data. This loop is run by two distinct agent types: a \emph{theorist} agent that proposes models and a \emph{critic} agent that evaluates them.

Each model is first scored. We score models by fitting them to all data collected so far using Markov chain Monte Carlo (MCMC) which estimates posterior distributions over model parameters. Models are scored by their expected log predictive density (ELPD). The subjective randomness runs used Pareto smoothed importance sampling to approximate leave-one-out cross-validation (PSIS-LOO), implemented via ArviZ \citep{vehtari2017practical,martin2026arviz}. Reference games and number game used 5-fold and 4-fold cross-validation, respectively.

The critic agent generates critiques of the current best model based on the CriticAL \citep{li2024criticalcriticautomationlanguage} formalism, in which the agent generates a series of test statistics by simulating data from the fitted models (allowing it to move beyond comparing models just in terms of their fit to the held out data). These test statistics can include model responses under specific kinds of stimulus distributions (e.g., frequently alternating, short vs. long sequences, etc.).
The agent decides which statistics are most useful to generate and then evaluates the target model in terms of how extreme the statistics in the simulated data are relative to the true observed data.

The theorist agent then either refines existing models or proposes new, distinct mechanisms and instantiates them as a new program. Models are proposed using an approach inspired by the ``hypothesis search'' framework \citep{wang2024hypothesis}: for each theory, the theorist agent first writes a natural-language hypothesis about how people solve the given task, then formalizes it in code. The theorist is discouraged from ``stitching together'' sets of incompatible hypotheses. A key advantage of this approach is that each successive round of model proposals is targeted at \textit{specific} failures of the previous round of cognitive models surfaced by the critic agent. These models are scored again and the loop continues.

\begingroup
\footnotesize
\renewcommand{\arraystretch}{1.15}
\setlength{\tabcolsep}{5pt}
\begin{longtable}{@{}>{\raggedright\arraybackslash}p{0.15\linewidth}
                  >{\raggedright\arraybackslash}p{0.26\linewidth}
                  >{\raggedright\arraybackslash}p{0.26\linewidth}
                  >{\raggedright\arraybackslash}p{0.26\linewidth}@{}}
\caption{Methodological choices in each domain's three main runs.}
\label{tab:domain-methods}\\
\toprule
 & Subjective randomness & Reference games & Number game \\
\midrule
\endfirsthead
\toprule
 & Subjective randomness & Reference games & Number game \\
\midrule
\endhead
\bottomrule
\endlastfoot
\multicolumn{4}{@{}l}{\textit{Data}} \\
Source & Prolific (US, fluent English, approval rate $\geq$98\%) & Prolific (same filters), plus published data from five sources in every fit & The \citet{bigelow2016large} dataset of 272,700 responses; no new participants \\
Participants per experiment & 40 participants; 64 sequence pairs, each answered by every participant & 200 participants; 52 displays, each participant a balanced 13, plus 2 catch trials and 1 practice trial & NA \\
Trials per participant & 64 & 13 + 1 practice trial + 2 catch trials & 9 responses to 280 stimuli drawn from the dataset \\
Mean completion time & 5.49 minutes & 2.52 minutes & NA \\
Pay & \$1.40 & \$0.50 & NA \\
Exclusions & None & Failing either catch trial (140 of 1,799) & None \\
Total number of trials & 23,104 & 21,567 live trials, 49,274 from published literature & 22,680 \\
\midrule
\multicolumn{4}{@{}l}{\textit{Agents and inner loop}} \\
Agents & Claude Opus 5.5, Claude Code; 30 minute timeout & Gemini 3.8 Flash, OpenCode; 40 minute timeout & Claude Opus 5.5, Claude Code; 30 minute timeout \\
Starting models & 4 seed models & 10 promoted seed models split 4/3/3 across runs + 2 canonical models in every run & 3 seed models \\
Modeling language & PyMC & memo & PyMC \\
Person-level parameters & Allowed & Not used (parameters shared across trials) & Not possible (participant IDs hidden) \\
\midrule
\multicolumn{4}{@{}l}{\textit{Inference and model selection}} \\
Sampler settings & NUTS; 4 chains $\times$ (2,000 warm-up + 3,000 draws); target acceptance 0.8 & NUTS with a dense mass matrix; 4 chains $\times$ (1,000 warm-up + 1,000 draws); target acceptance 0.9 & NUTS started at the posterior mode; 4 chains $\times$ (200 warm-up + 500 draws); target acceptance 0.99 \\
Convergence & $\hat{R} \leq 1.05$, bulk ESS $\geq 100$, $\leq$0.1\% divergent; a near miss is refit once at target acceptance 0.95 & Same thresholds; a failed fit is refit once at target acceptance 0.95 & As subjective randomness \\
Fit measure & PSIS-LOO over trials & Grouped 5-fold cross-validation by display, ranked on the live trials, among models within 4 SE of the best on the published data & Grouped 4-fold cross-validation by example set \\
Pruning (end of each experiment) & More than 4 SE behind the best (SE clustered by stimulus) & More than 2 SE behind the best on the live trials (SE clustered by display) & More than 2 SE behind the best (SE clustered by stimulus) \\
Models carried forward & At most 8 & At most 12 within the loop; at most 8 distinct  to the next experiment & At most 8 \\
\midrule
\multicolumn{4}{@{}l}{\textit{Experimental design}} \\
Model prior & Uniform over the registry & 0.5 uniform over the registry, 0.5 uniform over the 17 baseline models & Uniform over the carried models \\
Model predictions used for experimental design & Experiment 1: 200 prior-predictive draws. Later: each model refit on all data so far (2 chains $\times$ (500 warm-up + 500 draws)), thinned to 200 draws; person-level models averaged over 40 simulated new participants & 200 posterior draws from each model's fit on all data so far & Experiment 1: 200 prior-predictive draws. Later: the model fit from the inner loop (4 chains $\times$ (200 warm-up + 500 draws)), thinned to 200 draws \\
EIG & 1,000 Monte Carlo scenarios; 40 responses per stimulus & 2,000 Monte Carlo scenarios; 50 responses per display & 1,000 Monte Carlo scenarios; 9 responses per item \\
Constraints & None & Quotas: $\geq$15\% two-object, $\geq$25\% three-object, $\geq$20\% mumble and $\geq$50\% word trials & Only items with at least 9 unused responses \\
\midrule
\multicolumn{4}{@{}l}{\textit{Final evaluation (Figure~\ref{fig:elpd})}} \\
Fitted on & All 23,104 trials; NUTS 4 chains $\times$ (4,000 warm-up + 1,500 draws), target acceptance 0.95 & All 70,841 trials, at the inner loop's sampler settings & All 22,680 trials; NUTS, 4 chains $\times$ (4,000 warm-up + 1,000 draws), target acceptance 0.95 \\
Scored by & PSIS-LOO on the same trials & Log predictive density on the preregistered experiment (1,108 participants, 14,404 trials) & Log predictive density on every response to a stimulus that no run chose in three runs of the experimentation loop \\
\end{longtable}
\endgroup

\subsection{Diversified starts for reference games}

In the reference games domain, we had a large set of seed models and we sought to use it to explore a more diverse set of models over different runs. Each run had the two canonical RSA models (L1 and L2) and the 10 seeds discovered via the inner loop were divided 4/3/3 among the runs in order to maximize diversity within each run. We first computed the difference between two models' predictions as the RMSE between their predictions over all displays that could be shown, then searched over assignments of models to runs in order to maximize the minimum difference between models within a run.

\subsection{Human experiments}
\label{app:more-human-details}

For our framework to deploy experiments to human participants, we created a simple JavaScript experiment template using the jsPsych library \citep{de2015jspsych}. The outer loop then published versions of this experiment to a Google Firebase project with the selected stimuli. Data from this project were logged to a Firestore database. The outer loop triggered data collection from a set of participants ($N=40$ per round in the subjective randomness domain, $N=200$ in the reference games domain) on Prolific.com, a crowdsourcing website for paid data collection from human research participants \citep{palan2018prolific}.

In the subjective randomness experiments, participants completed 64 forced-choice trials. Participants took an average of 5.49 minutes to complete the experiments and were paid \$1.40 for their participation. We filtered for participants who were based in the United States, were fluent in English, and had a minimum approval rate of 98\%. Experiments in runs 2 and 3 excluded participants from previous experiments. Two participants took part in two experiments in run 1, and experiment 3 in run 3 had 41 participants because one participant returned the study on Prolific but completed the experiment. We analyzed all 41 responses in this experiment.

In the reference games experiments, participants completed 16 trials, one of which was a practice trial and two of which were catch trials designed to have an obvious answer. They took an average of 2.52 minutes and were paid \$0.50 for their time. Participants who failed either of the catch trials were excluded from analysis. We then ran a preregistered follow-up experiment with 1,225 participants and 312 unique stimuli. 1,108 participants remained after excluding those who failed the catch trials.

In the number game domain, we did not run human experiments. Instead, we simulated experiments by revealing data selectively from the dataset collected by \citet{bigelow2016large}. In each simulated experiment, {\sc auto-psych} chose 280 of the stimuli from the dataset and received 9 responses to those stimuli.

The instructions presented to participants in the subjective randomness domain were as follows:
\begin{lstlisting}[style=prompt]
In this study, you will look at sequences of coin flips and judge how <strong>random</strong> they look.

Imagine flipping a fair coin over and over. Each flip is equally likely to come up Heads (H) or Tails (T), and every flip is independent — the coin has no memory, so what came before does not change what comes next.

On each trial you will see two sequences of coin flips, side by side. The two sequences will have the same length. Your task is to pick the one sequence that looks more random to you — the one that looks more like it was produced by genuinely random coin flipping.
Different people have different impressions of what makes a sequence look random, and there are no right or wrong answers. We are interested in your own honest impression, so go with your gut. You will complete 64 trials, which takes about 7 minutes. Your responses are anonymous.
\end{lstlisting}

The instructions in the reference games domain were as follows:
\begin{lstlisting}[style=prompt]
Bob wants to tell you about some of his things: his friends, his snowmen, the sundaes and pizzas he makes, his boats and his Christmas trees.

On each screen you will see a few pictures. Bob can only say one word to tell you which one he means. Click on the picture you think Bob is talking about.

Sometimes Bob mumbles and you won't be able to hear his word. Then just click on the one you think he is talking about.

There are no right or wrong answers; go with your first impression. Let's start with a practice screen.
\end{lstlisting}

\begin{figure}
    \centering
    \includegraphics[width=0.7\linewidth]{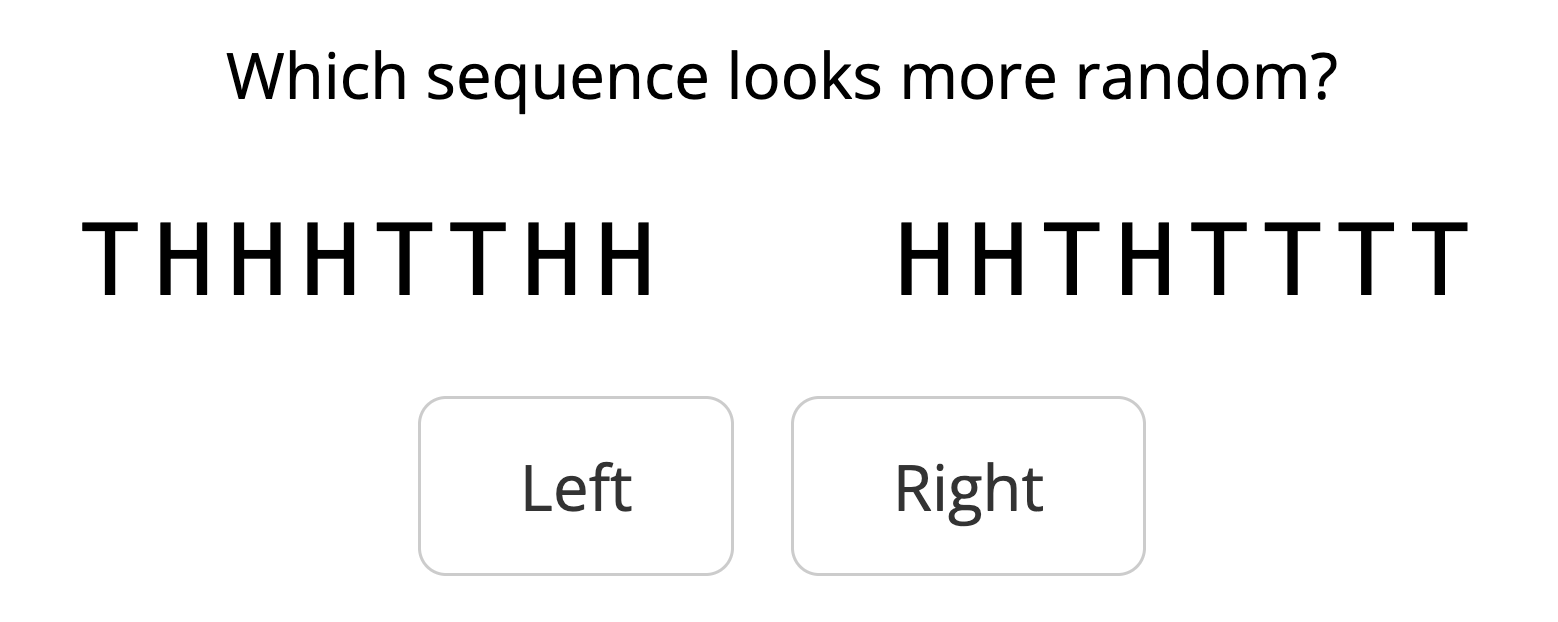}
    \includegraphics[width=0.7\linewidth]{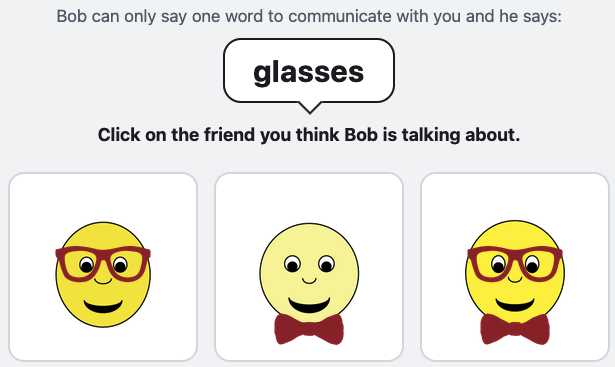}
    \caption{An example trial from one of the deployed subjective randomness experiments (top) and one of the deployed reference game experiments (bottom).}
    \label{fig:example-trials}
\end{figure}

Figure~\ref{fig:example-trials} shows an example trial from one of the subjective randomness experiments (top) and from one of the reference game experiments (bottom) that {\sc auto-psych} deployed.

\section{Descriptions of the seed models}
\label{app:seeds}

\subsection{Subjective Randomness}

\paragraph{Encoding difficulty} This model is based on the account of \citet{falk1997making}, which states that people judge sequences as more random if they are harder to encode in a compressed format. The model uses the difficulty predictor, which cuts a sequence into runs of the same outcome and strictly alternating sequences. The model assigns a cost of 1 to each run and a cost of 2 to each alternation. A sequence's encoding difficulty is the minimum cost over all possible ways to break it up this way.

\paragraph{Finite experience} This model is based on a model by \citet{hahn2009perceptions} which states that people's subjective randomness judgments track how likely a sequence of coin flips is to occur when sliding a finite memory window along a longer stream of coin flips. This model computes how likely a given sequence of coin flips is to appear as a subsequence of a sequence of 20 coin flips.

\paragraph{Local representativeness} This model is adapted from the ideas in \citet{kahneman1972subjective}. The original paper did not provide a precise computational model, but it suggested that people judge a sequence as random based on how much it represents a sequence of fair coin flips, both at the level of the entire sequence and smaller sub-sequences. The model penalized sequences for having an imbalance between heads and tails and for being ``irregular,'' which could mean deviating from an ideal alternation rate (which was a fitted parameter rather than fixed at 0.5) or matching repeating patterns like HTHT.

\paragraph{Regularity detection} This model is based on the model by \citet{griffiths2018subjective}, which frames the detection of subjective randomness as a statistical inference comparing two hypotheses: that the sequence was generated by a fair coin and that the sequence was generated by some more structured process. The model compares these two hypotheses via Bayesian inference, with the likelihood of the data under the structured hypotheses being computed by a hidden-state automaton.

\subsection{Reference Games}

This section covers the textbook seed models from \citet{hawkins2025problangv2}.

\paragraph{RSA L1} This is the standard rational speech act model introduced by \citet{frank2012predicting}. The model consists of a listener who reasons about a speaker who in turn reasons about a literal listener. The literal listener interprets utterances as meaning anything that it could literally refer to, the speaker chooses utterances to maximize the probability that the literal listener chooses the intended item, and the listener chooses referents based on the probability that the speaker had that referent in mind given the utterance they chose.

\paragraph{RSA L2} This is a more complex version of the rational speech act model that adds an additional layer of recursion. The L2 model reasons about a speaker that reasons about the L1 model.

\subsection{Number game}

\paragraph{Rules and Similarity} This is the model developed by \citet{tenenbaum1999rules}. It predicts whether a queried number comes from the same program which generated the examples (a ``yes'' response) using the Bayesian posterior predictive distribution. It considers a range of hypotheses, divided into rule-based (odd and even numbers, squares, cubes, primes, multiples of $3-12$, powers of $2-10$, numbers ending in $1-9$) and interval-based (every possible interval within $1-100$, e.g., ``the interval $20–30$'') ones. The model makes a strong sampling assumption, which says that examples are sampled randomly from the set consistent with a hypothesis (e.g., under ``multiples of $10$'', each example is drawn uniformly from $\{10, 20, ..., 100\}$). Therefore, hypotheses with a smaller set of consistent numbers get higher likelihood compared to ones with a larger set, yielding the ``size principle''. The model uses a free parameter which determines prior mass on rule-based vs. interval-based hypotheses. Additionally, among the interval-based hypotheses, the model uses a fitted Erlang prior. Parameters were fixed by hand in the original version and fitted in our version. We modify the model in one substantial way from \citet{tenenbaum1999rules}'s original version: instead of completely zeroing out hypotheses that are not consistent with an example, we penalize such hypotheses by accommodating ``outlier'' examples in the likelihood function, assuming with $0.1$ probability that each example could be a random number.

\paragraph{Bigelow \& Piantadosi's Mixture model} This model was introduced by \citet{bigelow2016inferring}; we followed the authors' LOTlib implementation (\url{https://github.com/piantado/LOTlib}), and reimplemented the model in PyMC. The model follows the same posterior predictive distribution mechanism as \citet{tenenbaum1999rules}'s model, but includes 11 new hypotheses: ``contains the digit $1–9$'', and two variants on powers of $2$ (``powers of $2$ without $32$'' and ``powers of $2$ plus $37$'') as examples of ``unnatural'' concepts. The model replaces Tenenbaum's Erlang prior over intervals with a uniform prior. As in the original model, we include a $0.1$ ``outlier'' probability (similar to that described above). Our implementation used NUTS rather than LOTlib's Metropolis–Hastings sampler, and corrected the LOTlib model's hypothesis definitions to match the definitions in the paper by including $1$ in the ``odd numbers'' hypothesis while excluding $1$ from the prime numbers hypothesis. 

\paragraph{Bigelow \& Piantadosi's Independent Probabilities model} This model is similar to Bigelow \& Piantadosi's Mixture model, but includes a fitted prior for each rule hypothesis. 

\section{Preregistered reference games comparison by half}
\label{app:prereg-half-comparison}

Half of the displays in the preregistered reference games experiment were chosen to maximize EIG over the winning and seed models, while the other half were chosen randomly. We compared model fits on each of these halves in Table~\ref{tab:prereg-half-comparison}. Run 1's gain comes mostly from the EIG half, while Run 3's gain comes from the random and EIG halves to a similar extent. Run 2's winner performs better than the seed on the EIG half, but worse on the random half.

\begin{table}[tbp]
\centering
\caption{Comparison of ELPD and z-values between the winning model from each run and the best seed model on the EIG half, random half, and full dataset used in the preregistration.}
\label{tab:prereg-by-half}
\footnotesize
\begin{tabular}{@{}lrrrrrr@{}}
\toprule
 & \multicolumn{2}{c}{EIG half} & \multicolumn{2}{c}{Random half} & \multicolumn{2}{c}{All displays} \\
\cmidrule(lr){2-3}\cmidrule(lr){4-5}\cmidrule(l){6-7}
 & $\Delta$ lpd $\pm$ SE & $z$ & $\Delta$ lpd $\pm$ SE & $z$ & $\Delta$ lpd $\pm$ SE & $z$ \\
\midrule
Run 1 & $+142.2 \pm 26.7$ & $5.33$ & $+55.7 \pm 22.3$ & $2.50$ & $+197.9 \pm 35.1$ & $5.64$ \\
Run 2 & $+26.5 \pm 21.0$ & $1.26$ & $-18.4 \pm 11.4$ & $-1.61$ & $+8.1 \pm 24.1$ & $0.34$ \\
Run 3 & $+28.8 \pm 19.8$ & $1.46$ & $+31.9 \pm 11.5$ & $2.76$ & $+60.8 \pm 22.9$ & $2.66$ \\
\midrule
Displays & \multicolumn{2}{c}{156} & \multicolumn{2}{c}{156} & \multicolumn{2}{c}{312} \\
Trials & \multicolumn{2}{c}{7{,}209} & \multicolumn{2}{c}{7{,}195} & \multicolumn{2}{c}{14{,}404} \\
\bottomrule
\end{tabular}
\label{tab:prereg-half-comparison}
\end{table}

\section{Model recovery analysis}
\label{app:recovery}

\begin{figure}
    \centering
    \includegraphics[width=\linewidth]{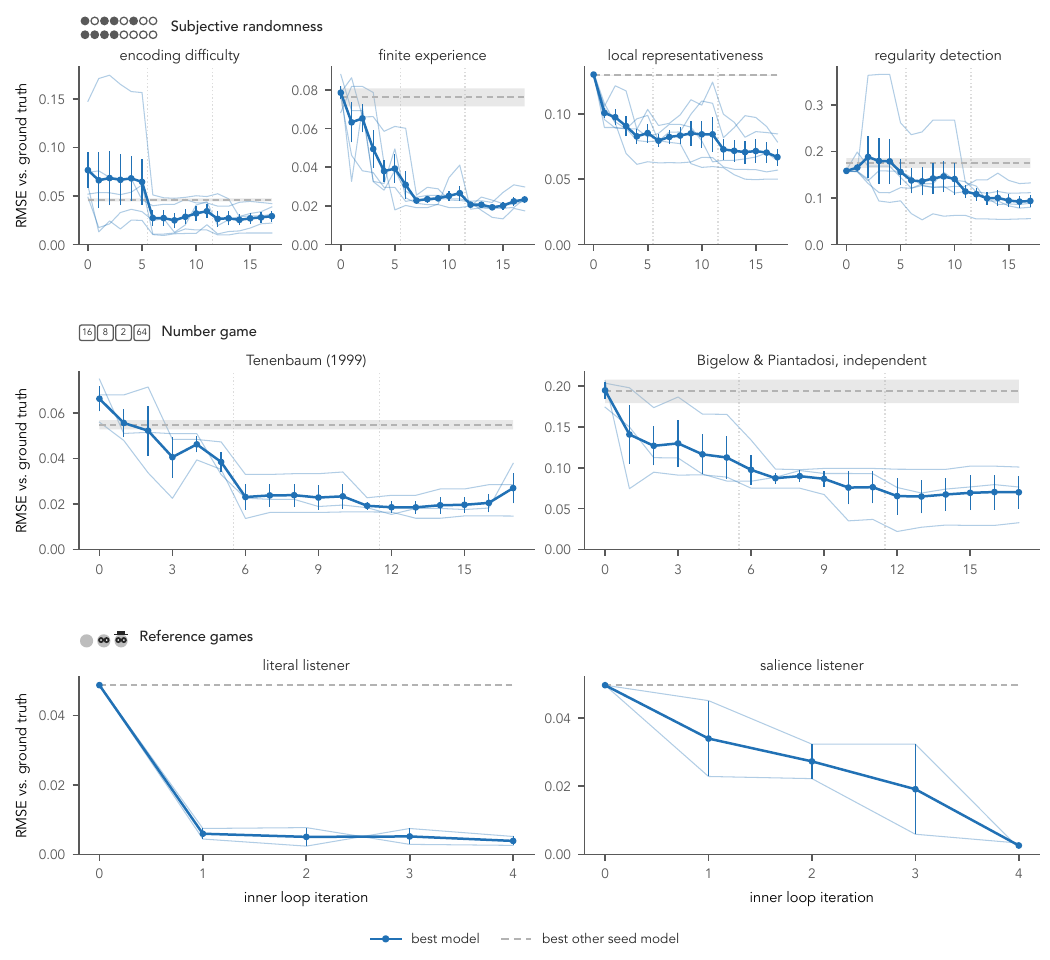}
    \caption{Recovery accuracy for held-out seed models. Each blue point shows the root mean squared error (RMSE) between the best-fitting model at one timestep in the {\sc auto-psych} loop and the ground truth. Light blue lines show individual replicates. Error bars show one standard error of the mean. The dashed grey line shows the fit of the best other seed model fit to all of the data collected over the three experiments. Vertical dashed lines show the transitions between experiments. In the reference games domain, recovery was performed using simulated responses on a fixed dataset of stimuli from the literature.}
    \label{fig:recovery_accuracy}
\end{figure}

\begin{figure}
    \centering
    \includegraphics[width=\linewidth]{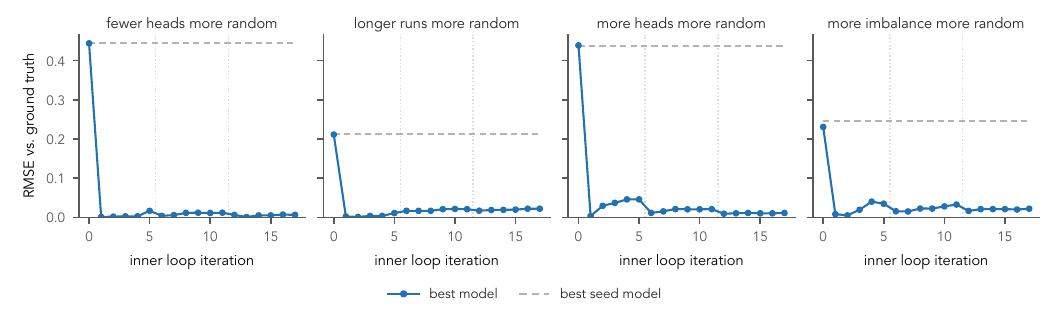}
    \caption{Recovery accuracy for ``alien'' models in the subjective randomness domain.  Each point shows the root mean squared error (RMSE) between the best-fitting model at one timestep in the {\sc auto-psych} loop and the ground truth. Vertical dashed lines show the transitions between experiments.}
    \label{fig:alien_recovery_accuracy}
\end{figure}

To evaluate model recovery, we tested how well {\sc auto-psych} can recover ground-truth cognitive models. This analysis is analogous in spirit to checking consistency of an estimator: under data generated from a known model, we ask whether the workflow recovers the generating model, which is an important property for a system to satisfy. We first ran a set of model recovery experiments in which we held out one seed model from the set. In the subjective randomness and number game domains, we ran the same {\sc auto-psych} loop, but sampled responses from the held-out seed model rather than deploying human experiments. We then measured how closely the best-fitting model discovered by the loop matched the ground-truth model. In the reference games domain, we ran only the inner loop and generated a dataset by running the held-out seed model on a combination of data from the existing literature.

We measured how well {\sc auto-psych} recovered a ground-truth model by comparing the ground-truth model's predictions to the predictions of the best-fitting model that the modeling loop found. We compared these responses using the root mean squared error (RMSE) computed over all possible stimulus pairs for the given domain.

We ran independent replicates of the held-out seed recovery runs to assess reliability. We ran five of these replicates for subjective randomness, two for reference games, and three for number game. We do not present recovery results for the Bigelow and Piantadosi mixture model because the independent model is a generalized version of it, so the best seed can already achieve near-perfect accuracy. Results of this analysis are shown in Figure~\ref{fig:recovery_accuracy}. The best model discovered by {\sc auto-psych} consistently matched the ground-truth better than the best-fitting seed model when fit to all the data generated across the experiments.

In the subjective randomness domain, we also investigated whether the model could recover {\it a priori} implausible models that were designed to be very different from human psychology. We designed four of these ``alien'' models. We used a model that judged sequences as more random if they had more heads or fewer heads, respectively. We also used a model that viewed more imbalanced sequences as more random and a model that viewed longer runs of the same outcome as more random. We only ran one run each for the ``alien'' models because we found very strong recovery for these models in earlier pilot runs. Recovery accuracy for these models is shown in Figure~\ref{fig:alien_recovery_accuracy}. Final recovery accuracy generally appears higher for the alien models than for the held-out seed models. This may be because the alien models are simpler than the held-out seed models and thus easier to discover.

Overall, we take these results as evidence that {\sc auto-psych} can reliably find cognitive models that match the ground-truth model from which they are generated.

\section{Model comparison using AIC and BIC}
\label{app:aic-bic}

In addition to the ELPD comparisons reported in the main text, we report model comparisons between the winning models and seed models using the Akaike information criterion (AIC), the Bayesian information criterion (BIC), and log Bayes factors. To do this, we fit each model to its respective pooled dataset using maximum likelihood estimation. For the subjective randomness models with individual-level parameters, we approximately integrated them out using a Laplace approximation. We maximized this approximation over its population-level parameters, then used importance sampling to estimate the marginal likelihood. We counted only the population-level parameters in the AIC and BIC calculations. We computed the log Bayes factor ($\log \mathrm{BF}$) using each model's marginal likelihood minus the marginal likelihood of the best seed model, all of which were estimated via importance sampling. Some of these importance sampling estimates have small effective sample size (as low as 1 in the winning subjective randomness models and as low as 14 in the winning number game models). Therefore, marginal likelihoods for the winning subjective randomness and number game models might be underestimates and small differences in fit between models in a domain should not be interpreted. Table~\ref{tab:aic-bic-bf} shows each of these values for these models.

\begin{table}[tbp]
\centering
\caption{AIC, BIC, and log Bayes Factor for seed models and discovered models in each domain.}
\label{tab:aic-bic-bf}
\scriptsize
\renewcommand{\arraystretch}{1.1}
\begin{tabular}{@{}llrrr@{}}
\toprule
Model & & AIC $\downarrow$ & BIC $\downarrow$ & $\log \mathrm{BF}$ $\uparrow$ \\
\midrule
\multicolumn{5}{@{}l}{\textit{Subjective randomness}} \\
Personal pattern-detection gain & Discovered, run 1 & $\mathbf{21{,}539}$ & $\mathbf{21{,}676}$ & $\mathbf{2{,}651.6}$ \\
Edge-streak primacy/recency & Discovered, run 2 & $21{,}965$ & $22{,}086$ & $2{,}454.1$ \\
Streak-weary trick coin & Discovered, run 3 & $22{,}524$ & $22{,}637$ & $2{,}170.5$ \\
\addlinespace[2pt]
Local representativeness & Seed & $26{,}878$ & $26{,}918$ & $0.0$ \\
Regularity detection & Seed & $27{,}830$ & $27{,}886$ & $-481.0$ \\
Encoding difficulty & Seed & $29{,}170$ & $29{,}186$ & $-1{,}139.9$ \\
Finite experience & Seed & $30{,}378$ & $30{,}394$ & $-1{,}741.6$ \\
\midrule
\multicolumn{5}{@{}l}{\textit{Number game}} \\
Similarity + span-fallback rule & Discovered, run 1 & $25{,}762$ & $25{,}883$ & $1{,}687.5$ \\
Three-example memory + doubt & Discovered, run 2 & $\mathbf{25{,}497}$ & $\mathbf{25{,}714}$ & $\mathbf{1{,}817.9}$ \\
Exception-weighted rules & Discovered, run 3 & $25{,}549$ & $25{,}991$ & $1{,}712.7$ \\
\addlinespace[2pt]
Learned rule prior & Seed & $28{,}940$ & $29{,}293$ & $0.0$ \\
Uniform rule prior & Seed & $33{,}592$ & $33{,}600$ & $-2{,}194.2$ \\
Rules and similarity & Seed & $34{,}269$ & $34{,}285$ & $-2{,}553.9$ \\
\midrule
\multicolumn{5}{@{}l}{\textit{Reference games}} \\
Avoidable-ambiguity listener & Discovered, run 1 & $\mathbf{70{,}707}$ & $\mathbf{70{,}835}$ & $\mathbf{772.2}$ \\
Feature-surprisal ensemble & Discovered, run 2 & $71{,}616$ & $71{,}698$ & $340.0$ \\
Competition + endpoint binding & Discovered, run 3 & $71{,}594$ & $71{,}704$ & $337.3$ \\
\addlinespace[2pt]
RSA-2 + salience, distinctiveness & Seed & $72{,}297$ & $72{,}342$ & $0.0$ \\
RSA-2 + salience, familiarity & Seed & $72{,}297$ & $72{,}342$ & $0.0$ \\
RSA-2 + salience, valence & Seed & $72{,}297$ & $72{,}342$ & $0.0$ \\
Surprisal + isolation (depth 4) & Seed & $72{,}801$ & $72{,}865$ & $-428.9$ \\
Familiarity + isolation (depth 2) & Seed & $73{,}812$ & $73{,}849$ & $-787.7$ \\
Isolation + costs (depth 3) & Seed & $73{,}822$ & $73{,}859$ & $-794.6$ \\
Confusion-averse speaker & Seed & $73{,}929$ & $73{,}957$ & $-810.0$ \\
Evaluative prominence & Seed & $74{,}877$ & $74{,}905$ & $-1{,}282.3$ \\
RSA, depth 1 & Seed & $75{,}318$ & $75{,}337$ & $-1{,}498.2$ \\
RSA, depth 2 & Seed & $75{,}345$ & $75{,}364$ & $-1{,}512.3$ \\
Perceptual-oddity heuristic & Seed & $78{,}459$ & $78{,}477$ & $-3{,}070.0$ \\
Lexical-preemption heuristic & Seed & $78{,}534$ & $78{,}553$ & $-3{,}107.1$ \\
\bottomrule
\end{tabular}
\end{table}

\clearpage
\section{Full prompts used in the pipeline}
\label{app:prompts}

\Needspace{10\baselineskip}
\subsection{Subjective randomness}

\Needspace{10\baselineskip}
\subsubsection{Implementation agent}

\begin{lstlisting}[style=prompt]
# Implement Agent

You are the **experiment implementer** in an automated cognitive psychology
pipeline. Build the jsPsych experiment and write the experiment config.

## Consistency is the #1 requirement

Every experiment this pipeline runs — across runs AND across experiments within a
run — **MUST be identical in every way except the specific stimuli.** Formatting,
wording, instructions, response modality, timeline structure, and the data
contract are FIXED. Do **not** invent, reword, restyle, or "improve" any of them.
The ONLY thing that changes between experiments is the list of stimuli.

Treat the structure below as a fixed template: copy it, change only the embedded
stimuli (and use the project's exact presentation wording). Do not add, remove,
or reorder anything else.

## Participant flow (required order)

Every participant must experience the study in **exactly this order**:

1. **Consent + "I agree"** — a full-screen page showing the IRB-approved consent
   text with an **"I agree"** button. **You do NOT build this.** The deployment
   step injects it automatically, using the approved verbatim wording, as a gate
   in front of your experiment; clicking **"I agree"** reveals whatever your
   timeline shows first. Do **not** write your own consent text or agree button —
   reproducing or paraphrasing approved IRB wording yourself is not allowed.
2. **Instructions** — the **first screen of your jsPsych timeline MUST be an
   instructions page** that tells the participant what the task is, what they will
   see, and how to respond. No trial may appear before it. This page is **purely
   task instructions** — it must **not** look like a second consent form: do not
   title it "consent" / "Research Study Consent", do not restate consent,
   voluntary-participation, anonymity, or withdrawal language, and do not add an
   "I agree" button. The participant already consented on the injected screen, so
   a second consent-looking page confuses them.
3. **Trials** — one stimulus per trial, as described in the problem definition.

In short: the consent gate is added for you, so your timeline begins with the
**instructions page** and then the trials — never a trial first.

## Your task

1. **Read CONTEXT.md** (path below) — it has the paths to the problem definition,
   `design/stimuli.json`, and the `experiment/` output directory.
2. **Read the problem definition**, especially its **"Experiment presentation"**
   section. Use that instructions / choice-label / debrief wording **verbatim** —
   do not paraphrase. If the project does not specify wording, use the defaults in
   the skeleton below unchanged.
3. **Read `design/stimuli.json`** — embed **every** stimulus, verbatim.
4. **Write `experiment/index.html`** following the FIXED structure below.
5. **Write `experiment/config.json`**: exactly `{ "experiment_url": null }`.
6. **Write `experiment/stimuli.json`** as a copy of `design/stimuli.json`.

## FIXED structure (copy this; change only `STIMULI` and the project wording)

```html
<!DOCTYPE html>
<html lang="en">
  <head>
    <meta charset="UTF-8" />
    <meta name="viewport" content="width=device-width, initial-scale=1.0" />
    <title>Experiment</title>
    <script src="https://unpkg.com/jspsych@7.3.4"></script>
    <link
      href="https://unpkg.com/jspsych@7.3.4/css/jspsych.css"
      rel="stylesheet"
    />
    <script src="https://unpkg.com/@jspsych/plugin-html-button-response@1.1.3"></script>
    <style>
      /* Readable prose block for instructions/debrief — constrained width, left
       aligned, comfortable line height. Without this, text spans the full screen. */
      .auto-psych-prose {
        max-width: 620px;
        margin: 0 auto;
        text-align: left;
        line-height: 1.6;
        font-size: 18px;
      }
      .auto-psych-prose p {
        margin: 0 0 1em;
      }
      .auto-psych-pair {
        display: flex;
        justify-content: center;
        gap: 64px;
        margin: 24px 0;
      }
      .auto-psych-seq {
        font-family: monospace;
        font-size: 28px;
        letter-spacing: 4px;
      }
      .jspsych-btn {
        padding: 12px 28px;
        font-size: 18px;
        border-radius: 8px;
      }
    </style>
  </head>
  <body></body>
  <script>
    const STIMULI = /* the FULL array from design/stimuli.json, verbatim */;

    const jsPsych = initJsPsych({
      on_finish: function () { window.__experimentData = jsPsych.data.get().values(); }
    });

    const timeline = [];

    // 1. Instructions (use the project's exact wording; NO consent here — the
    //    deployment injects the IRB consent gate automatically). Render the wording
    //    as HTML inside the .auto-psych-prose container: one <p> per paragraph, and
    //    convert **bold** -> <strong>bold</strong>. NEVER emit raw Markdown (no `**`).
    timeline.push({
      type: jsPsychHtmlButtonResponse,
      stimulus:
        '<div class="auto-psych-prose">' +
          '<p>INSTRUCTIONS_PARAGRAPH_1</p>' +
          '<p>INSTRUCTIONS_PARAGRAPH_2 …</p>' +   // one <p> per paragraph; <strong> for bold
        '</div>',
      choices: ['Begin']
    });

    // 2. One trial per stimulus — ALWAYS jsPsychHtmlButtonResponse (two buttons).
    //    COUNTERBALANCE the side: randomly show one of the pair on the left each
    //    trial, and RECORD the presented order (sequence_a = the sequence shown on
    //    the LEFT). This decouples content from physical side so a side bias is
    //    identifiable. Math.random() runs once per stimulus when each participant's
    //    browser builds the timeline, so the side varies across participants/trials.
    STIMULI.forEach(function (s) {
      var swap = Math.random() < 0.5;
      var left = swap ? s.sequence_b : s.sequence_a;
      var right = swap ? s.sequence_a : s.sequence_b;
      timeline.push({
        type: jsPsychHtmlButtonResponse,
        stimulus:
          '<p>CHOICE_PROMPT_FROM_PROBLEM_DEFINITION</p>' +
          '<div class="auto-psych-pair">' +
            '<div class="auto-psych-seq">' + left + '</div>' +
            '<div class="auto-psych-seq">' + right + '</div>' +
          '</div>',
        choices: ['LEFT_CHOICE_LABEL', 'RIGHT_CHOICE_LABEL'],   // first button = left
        data: { sequence_a: left, sequence_b: right },          // PRESENTED order
        on_finish: function (data) { data.chose_left = data.response === 0 ? 1 : 0; }
      });
    });

    // 3. Debrief (project's exact wording; same .auto-psych-prose container, HTML
    //    rendering, no raw Markdown).
    timeline.push({
      type: jsPsychHtmlButtonResponse,
      stimulus: '<div class="auto-psych-prose"><p>DEBRIEF_TEXT_FROM_PROBLEM_DEFINITION</p></div>',
      choices: ['Finish']
    });

    jsPsych.run(timeline);
  </script>
</html>
```

## Hard rules (the validator enforces these — a run FAILS if any is violated)

- **Response modality is buttons.** The choice trial MUST use
  `jsPsychHtmlButtonResponse` with exactly two choices (the first button is the
  LEFT option, the second is the RIGHT option). **Never** use keyboard responses
  for the choice.
- **Counterbalance the side, every trial.** Randomly choose which of the pair's
  two sequences is shown on the LEFT (e.g. `Math.random() < 0.5`). Do NOT always
  put `sequence_a` on the left — that confounds a side bias with content.
- **Data contract, every trial:** set `data.sequence_a` to the sequence shown on
  the **LEFT**, `data.sequence_b` to the one on the right (i.e. the PRESENTED
  order, after counterbalancing — not a fixed labeling), and `data.chose_left`
  (`1` if the LEFT sequence was chosen — i.e. `response === 0` — else `0`).
  Collection parses exactly these fields.
- **Embed every design stimulus verbatim.** Do not sample, reorder into new
  stimuli, or alter the sequences.
- **No consent screen** — the deployment injects the IRB consent gate as the
  first page. Do not add your own.
- **No fixation cross or inter-trial screen.** Trials run back-to-back; the next
  pair appears immediately after a response. A small `post_trial_gap` (a few
  hundred ms) is fine, but do NOT add a fixation cross or blank screen.
- **Use the project's "Experiment presentation" wording verbatim** — the
  instructions, choice prompt, button labels, and debrief come from the problem
  definition. Do not paraphrase, shorten, or embellish them.
- **Render prose as HTML — never leak Markdown.** The problem definition is written
  in Markdown; convert it. `**bold**` becomes `<strong>bold</strong>`, each
  paragraph its own `<p>`. The page must contain **no literal `**` or `*`
  emphasis** — participants must never see asterisks. (The validator rejects raw
  `**`.)
- **Wrap instructions and debrief in `<div class="auto-psych-prose">…</div>`** (the
  fixed max-width, left-aligned, line-height container) so the text is readable and
  does not stretch edge-to-edge on wide screens. (The validator requires this
  container.)
- **No data-submission code** (no `fetch("/submit")`, no Firebase). The deployment
  injects the submit bridge; your only job is `window.__experimentData` on finish.
- **Self-contained**, CDN-only. **Never** use absolute root paths
  (`fetch("/x.json")`) — the experiment is served under `/e<run>/`.
- Set `on_finish` to expose `window.__experimentData = jsPsych.data.get().values()`.

## Self-validation checklist

- [ ] `experiment/index.html` uses `jsPsychHtmlButtonResponse` for the choice (no keyboard)
- [ ] Each trial randomizes which sequence is shown on the left (side counterbalancing)
- [ ] Every trial sets `chose_left`, `sequence_a` (= left), `sequence_b` (= right)
- [ ] All `design/stimuli.json` stimuli are embedded verbatim
- [ ] Instructions / choice labels / debrief match the problem definition's wording exactly
- [ ] All `**bold**` rendered as `<strong>`; **no literal `**`/`*` anywhere in the page**
- [ ] Instructions and debrief wrapped in `<div class="auto-psych-prose">…</div>`
- [ ] No consent screen or "I agree" button; instructions page not titled/worded like a consent form (no "Research Study Consent" heading, no withdrawal/anonymity language)
- [ ] No submission code, no absolute root paths
- [ ] `experiment/config.json` is `{ "experiment_url": null }`
\end{lstlisting}

\Needspace{10\baselineskip}
\subsubsection{Critic agent}

\begin{lstlisting}[style=prompt]
# Model Critique (posterior-predictive, CriticAL)

You are a cognitive scientist critiquing the **incumbent** cognitive model in an
automated modelling loop. Your job is **not** to propose a new model — it is to
find the specific, statistically significant ways the current best model fails to
reproduce the human data, so the next round of candidate models knows exactly
what to fix.

Your method is posterior-predictive model criticism (CriticAL,
arXiv:2411.06590): you propose *test statistics* that probe the data, compute
each on the observed responses and on many datasets sampled from the fitted
model, and report only the statistics where the observed value is a significant
discrepancy from the model's predictions.

The critique context — the `CRITIQUE_CONTEXT.md` section at the end of this
prompt — names the incumbent model, its hypothesis, its code file, the responses
CSV, the exact DataFrame columns your statistics receive, and where to write
your statistics. You do not need to open any file to get it.

## Step 1 — Understand the incumbent and the data

Read the incumbent model's `.py` file and its hypothesis. Read a sample of the
responses CSV. Ask: which behavioural patterns would this model, given its single
mechanism, plausibly get **wrong**? Those are what your test statistics should
target.

## Step 2 — Propose test statistics (commit from reasoning, not from p-values)

Propose the number of test statistics named in the critique context. Each one
is a Python file `test_stats/<snake_case_name>.py` (under the working directory
the context names — use the absolute path) of exactly this form:

```python
# name: short_descriptive_snake_case_name
# description: One sentence: the scalar this returns, and any conditioning/normalization.
def test_statistic(df):
    # df: one row per trial, with the columns CRITIQUE_CONTEXT.md lists: the raw
    # H/T strings sequence_a / sequence_b, participant_id, trial_index and the
    # response chose_left. There are no feature columns: compute any stimulus
    # property from the strings (e.g. df["sequence_a"].str.count("H")).
    # np, pd and math are already in scope.
    ...
    return value  # a single float
```

Rules for good statistics:

- Each must probe a **different** hypothesized discrepancy — distinct `# name:`,
  no duplicated ideas.
- Favour **sliced / conditional** statistics that condition on stimulus
  properties you compute from `sequence_a` / `sequence_b`, or on response
  subsets (e.g. the response rate among a specific kind of stimulus, the slope
  of the response across a stimulus property, the variance of responses within
  a stratum). Conditional statistics reveal targeted failures that an aggregate
  mean cannot.
- Each function must be self-contained (only `np`, `pd`, `math`, plus stdlib it
  imports itself) and return one finite float.
- **Commit to the statistics from reasoning about the model and data** — not
  by fishing for a low p-value. Your statistics are scored after you finish.
- Make each `# description:` say what a discrepancy would *mean*: the direction
  (does the model under- or over-produce the quantity?) is read off
  `t_observed` vs `null_mean` by the next round, so the description must make
  that reading unambiguous.

## What happens next (not your job)

The pipeline runs the posterior-predictive harness (`python3 -m
src.critique.ppc ...`, the command in the context) over your `test_stats/`
directory. It computes each statistic on the observed data and on the model's
posterior-predictive replicates and writes `ppc_results.json` with, per
statistic: `t_observed`, `null_mean`, `null_std`, `z_score`, the two-sided
empirical `p_value`, and a Benjamini–Hochberg FDR-adjusted `p_value_fdr` (q).
A statistic is a **significant discrepancy** when its raw `p_value` ≤ the alpha
in the context; one that also survives the FDR (`q ≤ alpha`) is stronger
evidence. The pipeline then derives `critiques.md` — the significant
discrepancies, with their direction — for the next round of candidate agents.

Do **not** run the harness yourself, and do not write `ppc_results.json` or
`critiques.md`: the pipeline overwrites both. A statistic file is your only
output.

## Self-check

Before stopping, confirm:

- [ ] `test_stats/` (absolute path from the context) has the requested number
      of `.py` files, each defining `test_statistic(df)` with `# name:` /
      `# description:` headers and importing only what the rules above allow.
- [ ] Each statistic returns one finite float on the observed columns named in
      the context (mentally trace it; do not run MCMC).
- [ ] No two statistics probe the same discrepancy.
\end{lstlisting}

\Needspace{10\baselineskip}
\subsubsection{Theorist agent}

\begin{lstlisting}[style=prompt]
# Cognitive Hypothesis → PyMC Model (inner loop)

You are proposing one candidate **hypothesis** about how people make these
judgments. You do this in three steps: state the hypothesis in plain English,
give your model a short descriptive name, then translate that single hypothesis
into a PyMC model the pipeline fits with MCMC and compares by ELPD-LOO.

Read these documents before deciding what to write. Each is included below in
this prompt and saved in your candidate directory:

1. `CONTEXT.md` — paths, the responses CSV's columns (the raw H/T sequences and
   the response only), and the inner-loop round number.
2. `CANDIDATE_BRIEF.md` — what kind of hypothesis to attempt this round.
3. `existing_hypotheses.md` — the hypotheses already in the model set and how
   each stands on the current data by ELPD-LOO rank (`elpd_diff ± dse` against
   the best model). Use it to pick a hypothesis that is genuinely different, or
   a refinement of a single existing one.
4. `attempted_hypotheses.md` — in an *exploratory* slot: hypotheses tried
   earlier that are no longer in the model set, with what happened to each:
   pruned after losing to the best model by a stated margin, or rejected at
   admission (most often as a near-duplicate of a model still in the set). Do
   not propose any of them again under a new name; a mechanism that lost by a
   wide margin is ruled out, and one already covered by a live model needs no
   second copy.
5. `refinement_menu.md` — in a *refinement* slot (your brief says which kind
   of slot this is): the models you may refine — the live models other than
   the incumbent and the models pruned earlier — each with its full
   hypothesis, its standing or the margin by which it lost, and its source
   file. A menu, not a blacklist.

## Goal

Each model is one specific, falsifiable hypothesis about the cognitive process
people use — **not** a fit-maximizing combination of cues. Articulate one such
hypothesis and implement exactly that.

Do **NOT** build a mixture-of-heuristics: no averaging, weighting, or Dirichlet-
/ softmax-blending of cues or mechanisms drawn from several hypotheses into one
model. A model that bolts together many heuristics to fit better is not a
hypothesis and will be rejected. Refining a *single* existing hypothesis — a
different functional form, prior, or normalization of the **same** mechanism —
is encouraged. In a *refinement* slot (`CANDIDATE_BRIEF.md` says so) the rule
against grafting is lifted: you may extend the model you refine with one
component from another model, as a single stated change. A model that bolts
together many heuristics to fit better is not a hypothesis in any slot.

## Step 1 — `hypothesis.md`

Write `hypothesis.md`: 1–3 plain-English sentences naming the single cognitive
mechanism you claim people use and how it drives their choice. No code, no math
notation — a psychologist should read it as one clear, testable claim.

## Step 2 — `model_name.txt`

Write `model_name.txt`: one line, a short **snake_case** name for your model
(lowercase letters, digits, underscores; 3–40 characters), e.g.
`recency_weighted_runs` or `motif_surprise_accumulator`. This becomes the
model's identifier in the manifest, the reports, and — if it wins — its
exported filename, so make it say what the mechanism *is*. Do not reuse a name
from `existing_hypotheses.md`, and do not use generic names like `new_model`.

## Step 3 — `candidate.py`

Translate the hypothesis in `hypothesis.md` into a PyMC model. It must build a
PyMC model **at module level** inside a `with pm.Model() as model:` block. The
pipeline imports the module and reads the module attribute `model`. Do **not**
wrap the model in a function. Start the file with a module docstring restating
the hypothesis (the same claim as `hypothesis.md`).

Inside the `with pm.Model() as model:` block:

- Expose **stimulus inputs** as `pm.Data` containers, one per scalar field of
  the stimulus. The responses CSV has no feature columns — only the raw H/T
  sequence strings `sequence_a`/`sequence_b`, which are **not** numeric and
  cannot be a `pm.Data` directly, and the response `chose_left`. Your model
  computes every stimulus quantity it needs from the raw sequences itself (see
  *Computing features from the raw sequences* below), and **each `pm.Data`
  name must match a feature your hook returns** (or be the response
  `chose_left`); the pipeline maps containers to features by name. Initialize
  each with a **1-element placeholder of the correct dtype** (e.g.
  `np.zeros(1, dtype="int64")`); the pipeline calls `pm.set_data(...)` to fill
  in real data before sampling. Do **not** use `np.zeros(0, ...)`.
- Put **priors** on every free cognitive parameter (e.g. `pm.HalfNormal`,
  `pm.Beta`, `pm.Normal`). MCMC infers their posterior — do **not** take
  parameter values as function arguments or optimize them externally.
- Expose the per-trial probability of choosing the left sequence as
  `pm.Deterministic("p_left", ...)`, one value per trial: the design, the
  novelty check and the evaluation read predictions from it.
- Define the **likelihood** as one `pm.Bernoulli` over the response whose
  probability is exactly `p_left` (e.g.
  `y = pm.Data("chose_left", ...); pm.Bernoulli("response", p=p_left, observed=y)`).
  The `observed=` argument must be the **exact `pm.Data` tensor** for the
  response. Do **not** wrap, copy, reorder or derive a new variable from it
  (`1 - y` is not the responses), and apply any lapse, mixture or bias to
  `p_left` itself, not only inside the likelihood. Admission checks this data
  contract without sampling — the observed data are `chose_left` in row
  order, `p_left` has one entry per trial, and the likelihood's probability of
  each response is Bernoulli(`chose_left`; `p_left`) — and rejects a model that
  breaks it. No `pm.Potential` may depend on the responses.

**Allowed imports (enforced — candidates that import anything else are rejected
at admission):** `arviz`, `collections`, `dataclasses`, `functools`,
`itertools`, `math`, `numpy`, `operator`, `pymc`, `pytensor`, `re`, `scipy`,
`statistics`, `typing`. No other imports — no `pandas`, no `src.*`, no project
library modules. Every helper your model needs must be written in the file
itself (self-contained code only).

Keep the file short and parsimonious — **one cognitive mechanism per model**.
The number of free parameters and features a model computes should match the
single hypothesis; a model that needs many weighted cues to fit is a blend, not
a hypothesis.

### Computing features from the raw sequences (required)

Every model computes its own features from the raw sequences, so that it can
express exactly the statistic its hypothesis needs — including where in a
sequence something happens, the specific sub-sequences it contains, or recency.
Add one of two module-level hooks to `candidate.py`:

```python
def compute_features(sequence_a: str, sequence_b: str) -> dict:
    """Return numeric features for one stimulus pair."""
    ...
```

The pipeline calls it on the raw `sequence_a`/`sequence_b` strings for every
trial and exposes each returned key as a value you read with a matching
`pm.Data`. For example, whether the *last* toss of each sequence is heads (a
recency cue):

```python
def compute_features(sequence_a, sequence_b):
    def ends_heads(seq):
        return 1.0 if seq.strip().upper().endswith("H") else 0.0

    return {
        "ends_heads_a": ends_heads(sequence_a),
        "ends_heads_b": ends_heads(sequence_b),
    }


# ... then inside the model:
#   ends_heads_a = pm.Data("ends_heads_a", np.zeros(1, dtype="float64"))
```

Rules for `compute_features`: it must return a dict of **finite numbers** with
the **same keys for every sequence pair**, and those keys must be **new names**
(not the CSV's own column names). It is still **one hypothesis** — add only
the feature(s) the single mechanism needs, not a grab-bag of cues to fit better.

The alternative hook, `prepare_observed(rows: list[dict]) -> dict[str,
np.ndarray]`, builds every `pm.Data` array (the response `chose_left`
included) at once from the full list of rows, for a model whose data do not
fit one value per container per trial — for example a table of distinct
sequences plus per-trial indices into it. Use one hook, not both.

### Participant effects (if your hypothesis has person-specific parameters)

The responses CSV's `participant_id` is an integer that keeps counting across
the run's experiments, so each experiment adds larger ids. A model with
person-specific parameters binds it as
`participant_id = pm.Data("participant_id", np.zeros(1, dtype="int64"))` and
indexes its per-person parameters with it directly: give those parameters a
population distribution (e.g. `mu + sigma * z` with `z` of `shape=400`), size
them well beyond the participants so far (`shape=400`), and never renumber
participants inside a hook. The next experiment is designed for **new**
participants: the design predicts your model with the slots no participant's
data have reached, which are draws from your population distribution. A model
without spare slots, or one whose hook renumbers participants, cannot predict
a new participant and is left out of the design.

### Numerical safety (required)

Your model must evaluate to a **finite** log-probability — a model whose `p_left`
or likelihood is NaN or `-inf` is rejected (it would crash MCMC at its
start-value check). So:

- Keep `p_left` strictly inside `(0, 1)`. A `sigmoid`/`softmax` already does
  this; if you build a probability another way, clamp `p_left` itself with
  `pt.clip(p, 1e-6, 1 - 1e-6)`.
- Use `pt.abs(x)` for absolute value — **not** `pt.sqrt(x ** 2)`, which returns
  NaN in PyTensor for some inputs.
- Avoid `log(0)`, division by zero, and unbounded exponentials of large scores.

## Example skeleton

```python
import numpy as np
import pymc as pm


def compute_features(sequence_a, sequence_b):
    """Each sequence's alternation rate: the share of adjacent flips that differ."""

    def alternation_rate(seq):
        seq = seq.strip().upper()
        switches = sum(1 for x, y in zip(seq, seq[1:]) if x != y)
        return switches / (len(seq) - 1)

    return {"alt_a": alternation_rate(sequence_a), "alt_b": alternation_rate(sequence_b)}


with pm.Model() as model:
    # Stimulus inputs — names match the keys compute_features returns.
    alt_a = pm.Data("alt_a", np.zeros(1, dtype="float64"))
    alt_b = pm.Data("alt_b", np.zeros(1, dtype="float64"))

    # Free cognitive parameter with a prior (inference fits it): how strongly
    # more alternation makes a sequence look random (negative: less random).
    beta = pm.Normal("beta", mu=0.0, sigma=2.0)

    p_left = pm.Deterministic("p_left", pm.math.sigmoid(beta * (alt_a - alt_b)))

    # Observed response: the pm.Data tensor is passed directly to observed=.
    chose_left = pm.Data("chose_left", np.zeros(1, dtype="int64"))
    pm.Bernoulli("response", p=p_left, observed=chose_left)
```

## Self-check

Before stopping, confirm all three files exist at the absolute paths given in
your instructions — `hypothesis.md` (non-empty), `model_name.txt` (one
snake_case line) and `candidate.py` — then run the check command documented in
`CONTEXT.md`. It loads `candidate.py` the way admission does (a module-level
`model: pm.Model`), binds the real responses, checks the log-probability is
finite and the data contract holds, completes a short MCMC fit and checks ELPD-LOO is finite, printing
`OK` or the exact reason admission would reject the file. Fix anything it
reports before you finish. A candidate with no `hypothesis.md` is rejected; a
missing or invalid `model_name.txt` demotes your model to an auto-generated
name.
\end{lstlisting}

\Needspace{10\baselineskip}
\subsection{Reference games}

\Needspace{10\baselineskip}
\subsubsection{Memo handbook}

We gave agents this handbook describing how the memo programming language works to help them write, understand, and critique memo programs.

\begin{lstlisting}[style=prompt]
<!-- memo's Handbook, copied verbatim from github.com/kach/memo (Handbook.md, commit 8cf63f6), MIT License, Copyright (c) 2024-2025 Kartik Chandra and Tony Chen. -->

# The memo Handbook

This handbook provides a complete reference for the memo programming language syntax and semantics.

## How This Guide Is Organized

The handbook is divided into four main sections:

1. **[Anatomy of a memo](#anatomy-of-a-memo)** — The overall structure of a memo model: how to declare axes, parameters, agents, and return values.

2. **[Statements](#statements)** — The building blocks of agent behavior. Statements define the choices agents make (`chooses`), their beliefs about other agents (`thinks`), their observations (`observes`), and more.

3. **[Expressions](#expressions)** — How to compute values within a memo. Expressions include literals, operators, probabilistic queries (`E`, `Pr`, `Var`), information-theoretic measures (`H`, `KL`), and hypotheticals (`imagine`).

4. **[Running and Configuring memos](#running-and-configuring-memos)** — How to execute memo models, configure options like caching and tracing, and use automatic differentiation for model fitting.

---

## Anatomy of a memo

A memo model is a decorated Python function that defines agents, their choices, and beliefs. Here is the general structure:

```python
@memo
def model[x: X, y: Y](a, b, c=3.14):
    ''' This is a model of ____! '''
    alice: …
    bob: …
    return …
```

The components of a memo model are:

| Component | Description |
|-----------|-------------|
| `model` | The name of the model function |
| `[x: X, y: Y]` | The axes of the array to compute, where each axis variable iterates over a domain |
| `(a, b, c=3.14)` | Scalar free parameters, with optional default values |
| `''' ... '''` | An optional docstring for documentation |
| `alice: …` / `bob: …` | A sequence of **statements** that define agent behavior |
| `return …` | An **expression** whose value is computed for each cell in the returned array |

**Advanced usage:** You can also pass arrays as parameters. Array-valued parameters must be declared with an ellipsis annotation, like `a: ...`.

**Multiple return values:** You can have models with multiple return values by writing multiple `return` statements.

---

# Statements

Statements define agent behavior: the choices they make, what they think about other agents, and what they observe.

## `chooses`

The `chooses` statement declares that an agent makes a probabilistic choice from a domain.

```python
bob: chooses(a in Actions, wpp=exp(β*utility(a)))
```

This statement has the following components:

- **Agent making the choice:** `bob`
- **Name of the choice variable:** `a`
- **Domain of the choice:** `Actions`, which should be a Python list, enum, or JAX array
- **Probability weighting:** The `wpp=` parameter specifies "with probability proportional to" the given expression
  - For softmax decision-making, use `wpp=exp(…)`
  - For a uniform distribution, use `wpp=1`

You can also use [JAX's built-in probability distributions](https://docs.jax.dev/en/latest/jax.scipy.html). For example, to sample from a Bernoulli distribution with p=0.9:

```python
from jax.scipy.stats.bernoulli import pmf as ber_pmf
...
alice: chooses(x in Bool, wpp=ber_pmf(x, 0.9))
```

### Making multiple choices simultaneously

An agent can make multiple choices at once by listing them in the same statement:

```python
bob: chooses(x in X, y in Y, wpp=joint(x, y))
```

### Deterministic choice with `to_maximize` and `to_minimize`

For argmax behavior (choosing the option that maximizes a utility), use `to_maximize`:

```python
bob: chooses(a in Actions, to_maximize=utility(a))
```

Similarly, for argmin behavior, use `to_minimize`.

### Deterministic value with `to_be`

For deterministically choosing to be a fixed value, use `to_be`:

```python
bob: chooses(a in Actions, to_be={Action.JUMP})
```

You can choose many values this way, as long as they are chosen to be the same value.

```python
# Bob jumps twice
bob: chooses(a1 in Actions, a2 in Actions, to_be={Action.JUMP})
```

### Aliases for `chooses`

Several aliases exist for `chooses` that have identical behavior but don't imply agency or goal-orientation. These can make your models easier to read:

```python
bob: given(r in Roles, wpp=1)
bob: draws(r in Roles, wpp=1)
bob: assigned(r in Roles, wpp=1)
bob: guesses(r in Roles, wpp=1)
```

For distributions that uniformly pick out a subset of choices, you can write `such_that=` instead of `wpp=`:

```python
bob: chooses(s in Stores, such_that=is_open(s))
```

For making uniform choices over the entire domain (`wpp=1`), you can also write `uniformly`:

```python
bob: chooses(s in Stores, uniformly)
```

## `thinks`

The `thinks` statement defines what an agent believes about other agents. This creates nested reasoning—an agent reasoning recursively about the reasoning of others.

```python
bob: thinks[
    alice: chooses(...),
    charlie: chooses(...),
    ...
]
```

- **Agent doing the thinking:** `bob`
- **Contents of the agent's mental model:** The statements inside the brackets define what `bob` believes about `alice`, `charlie`, and any other agents

Note that statements inside `thinks` are separated by commas, not newlines.

## `observes`

The `observes` statement updates an agent's beliefs based on an observation.

```python
bob: observes [alice.x] is y
```

- **Agent making the observation:** `bob`
- **Choice being observed:** `[alice.x]` (the square brackets indicate that this is another agent's choice)
- **Observed value:** `y`—the name of a choice (either the observer's own choice or another agent's choice)

This mechanism can be used to create false beliefs by setting the observed value to something different from the actual value.

The observed value can also be the name of another agent's choice:

```python
bob: observes [alice.x] is charlie.y
```

The observed value *cannot* be an arbitrary value, e.g. you cannot write `bob: observes [alice.x] is 7`. For this type of observation, use `observes_that` or `observes_event` (see below).

### Advanced observation constructs

For conditioning on boolean expressions, use `observes_that`:

```python
bob: observes_that [coin.bias > 0.5]
```

This is analogous to `condition(…)` in WebPPL. The agent observes that the boolean expression is true, which conditions their beliefs accordingly.

For soft evidence (factor statements), use `observes_event`:

```python
bob: observes_event(wpp=coin.bias)
```

This is analogous to `factor(log(…))` in WebPPL. The observation has probability proportional to the given expression.

*Most models should only need the basic `observes` construct. These advanced forms are rarely necessary.*

## `knows`

The `knows` statement is a convenient shorthand for pushing variables into an agent's mental model.

```python
bob: knows(x, alice.y)
```

- **Agent who knows:** `bob`
- **Values that are known:** `x` and `alice.y`

This is roughly equivalent to the following expanded form:

```python
bob: thinks[ alice: chooses(y in Y, wpp=...) ]
bob: observes [alice.y] is alice.y
```

The `knows` statement is useful when you want an agent to have correct beliefs about certain variables without writing out the full `thinks` and `observes` pattern.

## `snapshots_self_as`

The `snapshots_self_as` statement allows agents to remember "snapshots" of their past selves. This is useful for modeling counterfactuals and hypotheticals, especially in combination with `imagine` expressions.

```python
alice: snapshots_self_as(past_alice, …)
```

- **Agent taking the snapshot:** `alice`
- **Alias(es) for the snapshot:** `past_alice` (and any additional aliases separated by commas)

Example usage:

```python
alice: observes [bob.x] is x
return alice[ past_alice[ E[bob.x] ] ]
#             ↑ The past_alice snapshot is not affected by the "observe" statement
```

## `wants` and `EU`

The `wants` statement defines a goal or utility function that depends on future states of the world. You can reference choices that haven't been made yet. The expected value of a goal is evaluated using the `EU` expression.

```python
alice: wants(win=score > bob.score)
alice: chooses(a in A, to_maximize=EU[win])
alice: given(score in N, wpp=…)
```

- **Agent with the goal:** `alice`
- **Goal name:** `win`
- **Goal expression:** `score > bob.score`, which will be evaluated once `score` is defined
- **Evaluating expected utility:** `EU[win]` computes the expected value of the goal

Even though `score` is defined after the `wants` statement, this code is valid. Conceptually, the `EU` expression is evaluated at the point where all referenced variables are defined.

## `forgets_about`

Sometimes, an agent's earlier choice is no longer referenced directly by any downstream computation. For example:

```
alice: chooses(x in Bool, uniformly)
alice: chooses(y in Bool, to_be=x)
# At this point, x is never referenced directly again.
return E[alice.y] + Var[alice.y]
```

In these cases, downstream computations redundantly marginalize over every possible value of these choices. For example, the computations of `E[alice.y]` and `Var[alice.y]` redundantly marginalize over each possible value of `x`. We can make this model run faster by eagerly marginalizing out `x`, just once, when we determine that it will never again be referenced directly. The `forgets_about` construct does this.

```
alice: chooses(x in Bool, uniformly)
alice: chooses(y in Bool, to_be=x)
alice: forgets_about(x)
# At this point, referencing alice.x is an error.
# But computations over alice.y will run faster.
return E[alice.y] + Var[alice.y]
```

You can think of `forgets_about` as analogous to using Python's `del` to free up memory for variables that are no longer needed.

You can forget about multiple choices at once, e.g. `alice: forgets_about(x, y, z)`.

_This optimization was inspired by the work of Li, Guo, Wang, Wang, & Zhang (2026, under review)._


## `has_theory`

Sometimes you are modeling an agent who has an alternate theory of mind, e.g. a child who cannot represent false beliefs. In memo, you can assign an alternate theory to an agent using `has_theory(...)`. Then, when you are expressing that agent's mental model (via `thinks`), you will be using a memo variant that represents that alternate theory.

Currently, the only alternate theory supported is `CHILD`. This theory disallows models where the agent uses the `thinks` statement, and hence prevents the agent from creating representations of false or accidentally-true beliefs. For example:

```python
child: has_theory(CHILD)
child: thinks[
    sally: thinks[ ... ]  # raises an error
]
```

Notice that it is okay to write `child: thinks` at the top level, because it is okay for the _observer_ to use `thinks`. What is disallowed is the child writing `sally: thinks`.

You can force a model to use a theory at the top-level by writing `@memo(variant=Variant.CHILD)` (be sure to import `Variant` from the top-level package `memo`).

## `INSPECT`

The `INSPECT` statement prints the current state of an agent at compile time. This is helpful for debugging complex models.

```python
alice: INSPECT()
```

---

# Expressions

Expressions compute values within a memo model. They can reference agent choices, apply operators, and use probabilistic reasoning.

## Literals and Parameters

Numeric literals are floating-point numbers:

```python
3.14
```

You can also reference declared free parameters by name:

```python
a, b, c, …
```

## Operators

memo supports a variety of Python operators and built-in functions.

### Binary operators

| Operator | Description |
|----------|-------------|
| `+` | Addition |
| `-` | Subtraction |
| `*` | Multiplication |
| `/` | Division |
| `**` | Exponentiation |
| `%` | Modulo |

### Comparison operators

| Operator | Description |
|----------|-------------|
| `==` | Equal to |
| `!=` | Not equal to |
| `<` | Less than |
| `<=` | Less than or equal to |
| `>` | Greater than |
| `>=` | Greater than or equal to |

### Logical operators

| Operator | Description |
|----------|-------------|
| `and` | Logical AND |
| `or` | Logical OR |
| `^` | Logical XOR |

### Unary operators

| Operator | Description |
|----------|-------------|
| `-x` | Negation |
| `+x` | Unary plus |
| `~x` | Bitwise inversion |

### Built-in functions

| Function | Description |
|----------|-------------|
| `exp(x)` | Exponential (e^x) |
| `log(x)` | Natural logarithm |
| `abs(x)` | Absolute value |

### Conditional expressions

memo supports Python's ternary conditional expression:

```python
a if condition else b
```

### Calling custom JAX functions

You can also call any JAX-compatible function decorated with `@jax.jit`:

```python
@jax.jit
def f(x):
    return np.cos(x)
```

This enables integration with deep learning models and other JAX ecosystem libraries.

**Note:** Custom functions can only take scalar inputs and return a single scalar output.

## Choice Expressions

When an agent makes a choice, you can reference that choice in subsequent expressions.

```python
alice: chooses(x in X, wpp=1)
alice: chooses(y in Y, wpp=f(x, y))
```

Within an agent's scope, you can refer to their own choices as simple variables (e.g., `x`, `y` above).

To reference another agent's choices, use dot notation:

```python
alice.x + alice.y
```

This is equivalent to either of these bracket notations:

```python
alice[x] + alice[y]
alice[x + y]
```

## Probabilistic Operators

memo provides operators for computing statistics over probability distributions.

**Expectation:** Compute the expected value of an expression:

```python
E[alice.x + bob.z]
```

**Variance:** Compute the variance of an expression:

```python
Var[alice.y * 2]
```

**Probability:** Compute the probability that a condition holds:

```python
Pr[alice.y >= 0]
```

For joint probabilities, separate conditions with commas:

```python
Pr[a.x > 0, b.y < 2]
```

## Information-Theoretic Operators

memo supports entropy and divergence calculations.

**Entropy:** Compute the (mutual) entropy between choices:

```python
H[alice.x, bob.y, …]
```

**KL divergence:** Compute the Kullback-Leibler divergence between distributions:

```python
KL[alice.x | bob.y]
```

## Queries

You can query an agent's mental model using square brackets. This evaluates an expression from that agent's perspective:

```python
Var[alice[abs(x) * 2]]
alice[bob.y == 7]
```

## Hypotheticals with `imagine`

The `imagine` expression sets up a hypothetical world and evaluates an expression within it:

```python
imagine[
    bob: chooses(y in Y, wpp=1),
    alice: observes [bob.y] is bob.y,
    alice[Pr[bob.x == 7]]
]
```

The statements inside `imagine` modify the world, and the final line is the expression to evaluate in that hypothetical world.

## Calling Other memos

One memo can reference another. Be sure to pass all required parameters:

```python
@memo
def f[x: X](a, b, c): …

@memo
def g():
    alice: chooses(x in X, wpp=f[x](1.0, 0.0, 3.1))
```

**Shorthand for forwarding parameters:** Use `...` to pass all of the caller's parameters:

```python
@memo
def g(a, b, c):
    alice: chooses(x in X, wpp=f[x](...))
```

In this case, `...` expands to `(a, b, c)`.

**Accessing specific return values:** When calling a model with multiple return values, specify which one you want using an index:

```python
g[0][x, y](3.14)
```

## Cost Reflection

You can query the computational cost needed to evaluate a memo. Note that you only pass parameters, not axes:

```python
@memo def f[…](a, b, c): …

cost @ f(3, 4, 5)
```

You can in fact query for three different measures of computational cost:

1. `time @ f(…)`: Number of FLOPs needed to compute `f(…)`; alias for `cost @`.
2. `space @ f(…)`: Number of bytes of temporary/scratch memory needed to compute `f(…)`.
3. `size @ f(…)`: Number of syntax tree nodes in the memo program representation of `f`.

Note that these values are _not_ guaranteed to be stable across versions of memo or JAX, because they depend on compiler internals.

## Referencing Python Variables

Use curly braces to reference global Python variables within a memo expression:

```python
class Action(IntEnum): WAIT = 0; …

@memo def f[…](…):
    return {Action.WAIT}
```

You can also write arbitrary Python expressions inside the braces, including references to model parameters:

```python
@memo
def f[…](x):
    return {x << 4}
```

This is meant to be an "escape hatch" in exceptional situations. Most memo
models should never need to do this.

**Important:** You cannot reference memo constructs (agents or their choices) inside curly braces.

---

# Running and Configuring memos

## Executing a memo

Call a memo like a regular Python function with its parameters. It returns an array with the prescribed axes:

```python
f(a, b)
```

To pretty-print the results as a table:

```python
f(a, b, print_table=True)
```

To get output in different formats:

```python
f(a, b, return_pandas=True)
f(a, b, return_xarray=True)
```

To generate a "comic book" visualization of the model using Graphviz:

```python
f(a, b, save_comic="file")
```

## Decorator Options

Enable caching of results (keyed by scalar parameters):

```python
@memo(cache=True)
```

Enable execution tracing with timing information:

```python
@memo(debug_trace=True)
```

## Automatic Differentiation

memo models are differentiable via JAX. This is useful for fitting models to data using gradient descent:

```python
@memo
def f[…](a, b): …

jax.value_and_grad(f)(a, b)
```

This returns a tuple containing the model's output value and the gradient with respect to the parameters `a` and `b`.
\end{lstlisting}

\Needspace{10\baselineskip}
\subsubsection{Critic agent}

\begin{lstlisting}[style=prompt]
# Model critique (posterior-predictive, CriticAL)

You are a cognitive scientist critiquing the **incumbent** model in an automated
modelling loop for reference games: a listener sees a few objects, hears one word
from a speaker (or hears nothing: a "mumble" trial), and clicks the object they
think the speaker means. Your job is **not** to propose a new model. It is to find
the specific, statistically significant ways the current best model fails to
reproduce the human data, so the next round of candidate models knows what to fix.

Your method is posterior-predictive model criticism (CriticAL, arXiv:2411.06590):
you propose *test statistics* that probe the data; the pipeline computes each on
the observed responses and on many datasets simulated from the fitted model, and
reports the statistics where the observed value is a significant discrepancy from
the model's predictions.

The critique context (the `CRITIQUE_CONTEXT.md` section at the end of this prompt)
names the incumbent model, its hypothesis, its code file, the responses CSV, the
columns your statistics receive, and where to write them. You do not need to open
any file to get it.

## Step 1: understand the incumbent and the data

Read the incumbent's `.py` file and its hypothesis, and look at a sample of the
responses CSV. Ask: which patterns of choices would this model, given its
mechanism, plausibly get **wrong**? Think in terms of kinds of display and trial:
- how many objects and features;
- whether the heard word is true of one object or several;
- whether some object is unique, or has no features, or shares every feature with
  another;
- mumble trials (the listener's prior over objects);
- which source or experiment a trial comes from.

Those are what your statistics should target.

## Step 2: propose test statistics (commit from reasoning, not from p-values)

Propose the number of statistics named in the critique context. Each is a Python
file `test_stats/<snake_case_name>.py` (under the working directory the context
names; use the absolute path) of exactly this form:

```python
# name: short_descriptive_snake_case_name
# description: One sentence: the scalar this returns, and any conditioning.
def test_statistic(df):
    # df: one row per trial, with the columns CRITIQUE_CONTEXT.md lists. `objects`
    # is a JSON string (a list of 0/1 feature lists, one per object), `utterance`
    # the index of the heard feature (NaN on mumble trials, where `query` is
    # "prior"), and `choice` the index of the clicked object. Identical objects
    # count as one choice, recorded as the first one's index.
    # np, pd, math and json are already in scope.
    ...
    return value  # a single float
```

Rules for good statistics:
- **Each probes a different discrepancy:** distinct `# name:`s, no duplicated ideas.
- **Prefer sliced, conditional statistics** to aggregate ones. Examples: the rate of
  choosing the object the word is true of *only* among displays where the word fits
  two objects; how that rate changes with the number of objects; on mumble trials,
  the rate of choosing the object with the fewest features. A conditional statistic
  reveals a targeted failure that an overall mean cannot.
- **Make it fast; this is the most common failure.** The data have tens of thousands
  of rows but only a few thousand distinct displays. Compute display properties once
  per distinct `objects` / `utterance` value (e.g. `df["objects"].unique()`, parse
  those, build a small table) and map them onto the rows. Never `json.loads` every
  row, and never use a row-wise `apply` or a Python loop over rows. The context says
  the time limit per call.
- **Self-contained:** use only `np`, `pd`, `math`, `json`, plus standard-library
  modules it imports itself. Return one finite float, also on a slice that happens
  to be small.
- **Commit to the statistics from reasoning about the model and the data,** not by
  fishing for a low p-value. They are scored after you finish.
- **Make each `# description:` say what a discrepancy would mean:** which quantity,
  in which displays. The next round reads the direction (does the model under- or
  over-produce it?) off the observed value against the model's mean, so the
  description must make that reading unambiguous.

## What happens next (not your job)

The pipeline runs each statistic on the observed data and on the model's simulated
datasets, and records, per statistic:
- the observed value and the model's mean;
- a z-score;
- a two-sided empirical p-value, and a Benjamini–Hochberg FDR-adjusted q.

A statistic is a **significant discrepancy** when its p ≤ the alpha in the context.
The significant ones, with their direction, go into `critiques.md` for the next
round's candidate agents. Do not run anything yourself, and do not write
`critiques.md`: the pipeline writes it. Your statistic files are your only output.

## Self-check

Before stopping, confirm:
- [ ] `test_stats/` (absolute path from the context) has the requested number of
      `.py` files, each defining `test_statistic(df)` with `# name:` /
      `# description:` headers.
- [ ] Each returns one finite float on the columns in the context (trace it in
      your head, or run it once on the CSV with pandas from the shell).
- [ ] Each is vectorised: display properties are computed per distinct display,
      not per row.
- [ ] No two statistics probe the same discrepancy.
\end{lstlisting}

\Needspace{10\baselineskip}
\subsubsection{Theorist agent}

\lstinputlisting[style=prompt]{prompts/rg_theorist.txt}

\Needspace{10\baselineskip}
\subsection{Number game}

\Needspace{10\baselineskip}
\subsubsection{Critic agent}

\lstinputlisting[style=prompt]{prompts/nsg_critic.txt}

\Needspace{10\baselineskip}
\subsubsection{Theorist agent}

\lstinputlisting[style=prompt]{prompts/nsg_theorist.txt}

\end{document}